\documentclass[11pt,a4paper]{article}
\usepackage[hyperref]{emnlp2020}
\usepackage{times}
\usepackage{latexsym}
\usepackage{times}
\usepackage{algorithmic}
\usepackage{algorithm}
\usepackage{color}
\usepackage{booktabs}
\usepackage{xspace}
\usepackage{url}
\usepackage{mathtools}
\usepackage{amsmath}
\usepackage{float}
\usepackage{tabularx}
\usepackage{graphicx}
\usepackage{subcaption}
\usepackage{multirow}
\usepackage{enumerate}
\usepackage{esvect}
\usepackage{adjustbox}
\usepackage{verbatim}
\usepackage{amsmath}
\usepackage{indentfirst}
\usepackage{amssymb}
\usepackage{siunitx}
\usepackage{xcolor}
\usepackage{placeins}

\usepackage{url}
\usepackage{microtype}
\aclfinalcopy 
\def\aclpaperid{2297} 

\title{Neutralizing Gender Bias in Word Embeddings with \\ Latent Disentanglement and Counterfactual Generation}

\author{Seungjae Shin, Kyungwoo Song, JoonHo Jang, Hyemi Kim, Weonyoung Joo, Il-Chul Moon \\
  Korea Advanced Institute of Science and Technology (KAIST), Daejeon, Korea \\
  \texttt{\{tmdwo0910,gtshs2,adkto8093,khm0308,es345,icmoon\}@kaist.ac.kr} \\}

\date{}
\DeclareMathOperator{\Ked}{k}
\DeclareMathOperator{\K}{k}
\DeclareMathOperator{\Ker}{\boldsymbol{K}}

\begin{document}
\maketitle
\begin{abstract}
Recent research demonstrates that word embeddings, trained on the human-generated corpus, have strong gender biases in embedding spaces, and these biases can result in the discriminative results from the various downstream tasks. Whereas the previous methods project word embeddings into a linear subspace for debiasing, we introduce a \textit{Latent Disentanglement} method with a siamese auto-encoder structure with an adapted gradient reversal layer. Our structure enables the separation of the semantic latent information and gender latent information of given word into the disjoint latent dimensions. Afterwards, we introduce a \textit{Counterfactual Generation} to convert the gender information of words, so the original and the modified embeddings can produce a gender-neutralized word embedding after geometric alignment regularization, without loss of semantic information. From the various quantitative and qualitative debiasing experiments, our method shows to be better than existing debiasing methods in debiasing word embeddings. In addition, Our method shows the ability to preserve semantic information during debiasing by minimizing the semantic information losses for extrinsic NLP downstream tasks.
\end{abstract}

\section{Introduction}
\noindent Recent researches have disclosed that word embeddings contain unexpected bias in their geometry on the embedding space \cite{bolukbasi16,zhao2019gender}. The bias reflects unwanted stereotypes such as the correlation between gender\footnote{While we acknowledge a potential and expanded definition on gender as stated in \citet{larsongender}, we only cover the gender bias between the male and female in this paper.} and occupation words. \citet{bolukbasi16} enumerated that the automatically generated analogies of ($she,he$) in the Word2Vec \cite{mikolov2013distributed} show the gender biases in significant level. An example of the analogies is the relatively closer distance of \textit{she} to \textit{nurse}; and \textit{he} to \textit{doctor}.
\begin{figure}[t]
\centering
\includegraphics[width=\columnwidth,height=2.2in]{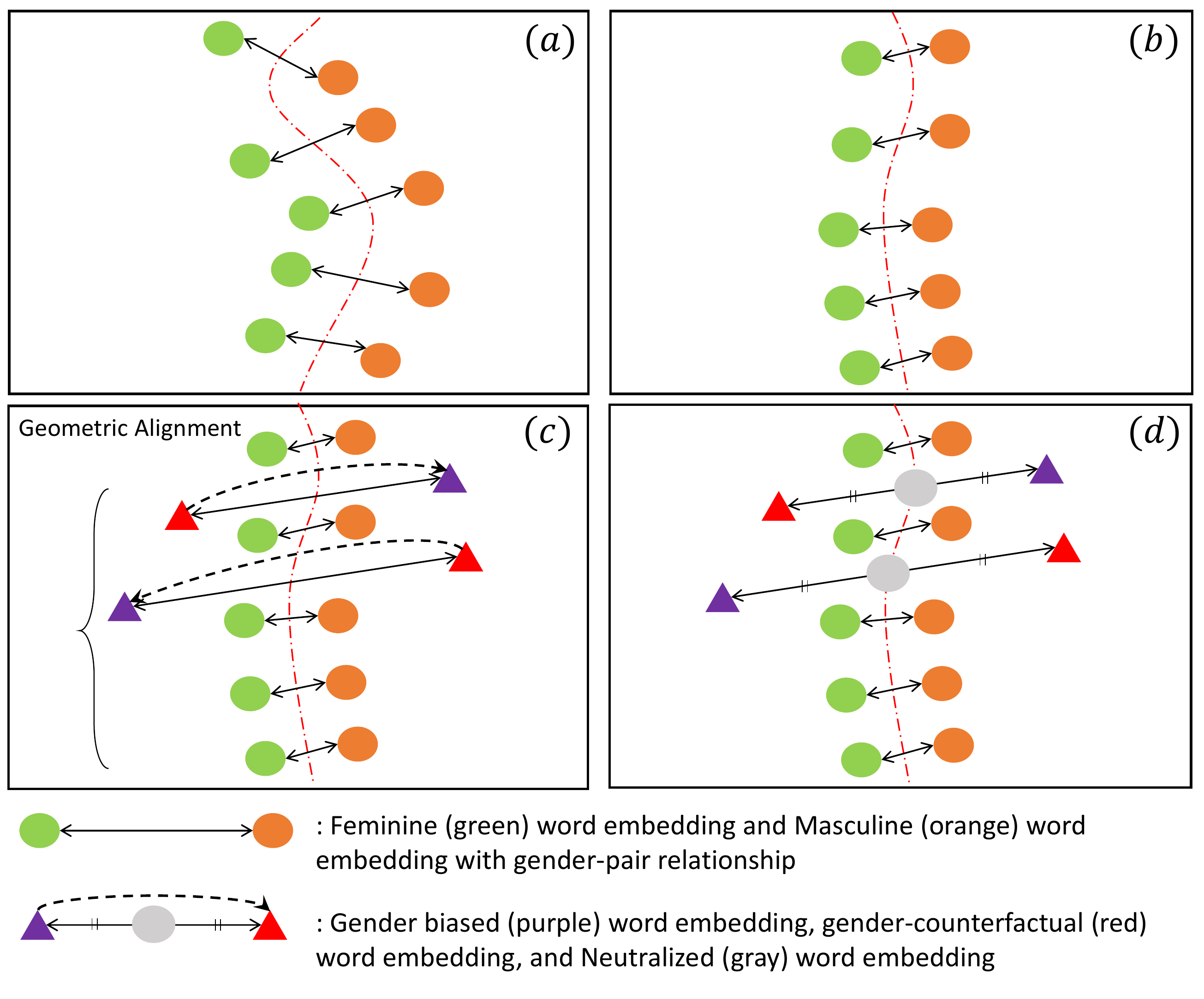}
\caption{The process view of our method. We can improve the embedding space from $(a)$ to $(b)$ with a better-aligned structure between gender word pairs by the proposed latent disentanglement. Afterwards, $(c)$ we generate the gender-counterfactual embedding of the gender-biased word while keeping a geometrically aligned relationship with the gender word pairs to guarantee that the pair of word embeddings only differs from gender information, not hurting semantic information. $(d)$ We obtain the gender-neutralized word embedding by interpolating the embedding from the pair of original-counterfactual word embeddings.
}\label{fig:schematic}
\end{figure}
\citet{garg18} demonstrated that the embeddings, from Word2Vec \cite{word2vec13} to Glove \cite{glove14}, have strong associations between value-neutral words and population-segment words, i.e. a strong association between \textit{housekeeper} and \textit{Hispanic}. This unwanted bias can cause biased results in the downstream tasks \cite{caliskan17,kiritchenko2018examining,bhaskaran2019good} and gender discrimination in NLP systems.

\indent From the various gender debiasing methods for pre-trained word embeddings, the widely recognized method is a post-processing method, which projects word embeddings to the space that is orthogonal to the gender direction vector defined by a set of gender word pairs. However, if the gender direction vector includes a component of semantic information\footnote{Throughout this paper, we define the semantics of words to be the meanings and functionality of words other than the gender information by following \citet{2019room}.}, the semantic information will be lost through the post-processing projections.

\indent To balance between the gender debiasing and the semantic information preserving, we propose an encoder-decoder framework that disentangles a latent space of a given word embedding into two encoded latent spaces: the first part is the gender latent space, and the second part is the semantic latent space that is independent to the gender information. To disentangle the latent space into two sub-spaces, we use a gradient reversal layer by prohibiting the inference on the gender latent information from the semantic information. Then, we generate a counterfactual word embedding by converting the encoded gender latent into the opposite gender. Afterwards, the original and the counterfactual word embeddings are geometrically interpreted to neutralize the gender information of given word embeddings, see Figure \ref{fig:schematic} for the illustration on our debiasing method.

Our contributions are summarized as follows:
\begin{itemize}
\item We propose a method for disentangling the latent information of the word embedding by utilizing the siamese auto-encoder structure with an adapted gradient reversal layer.
\item We propose a new gender debiasing method, which transforms the original word embedding into gender-neutral embedding, with the gender-counterfactual word embedding.
\item We propose a generalized alignment with a kernel function that enforces the embedding shift, during the debiasing process, in a direction that does not damage the semantics of word embedding.
\end{itemize}

\indent We evaluated the proposed method and other baseline methods with several quantitative and qualitative debiasing experiments, and we found that the proposed method shows significant improvements from the existing methods. Additionally, the results from several NLP downstream tasks show that our proposed method minimizes performance degradation than the existing methods.

\section{Gender Debiasing Mechanisms for Word Embeddings}
\noindent We can divide existing gender debiasing mechanisms for word embeddings into two categories. The first mechanism is neutralizing the gender aspect of word embeddings in the training procedure. \citet{zhao18} proposed the learning scheme to generate a gender-neutral version of Glove, called GN-Glove, which forces preserving the gender information in pre-specified embedding dimensions while other embedding dimensions are inferred to be gender-neutral. However, learning new word embeddings for large-scale corpus can be difficult and expensive.\\
\begin{figure*}[t]
	\centering
	\includegraphics[width=6.4in,height=2.9in]{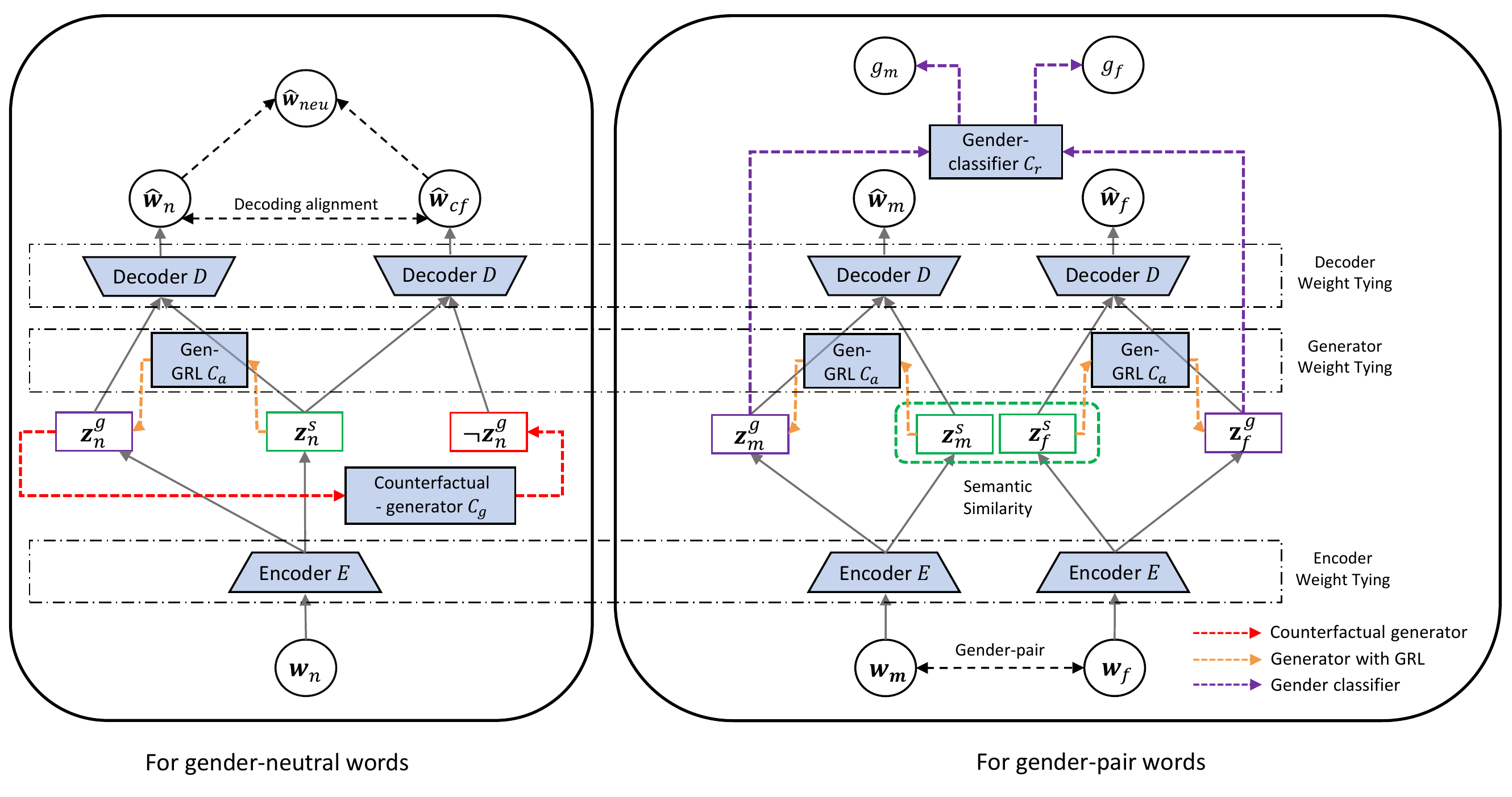}
	\caption{The framework overview of our proposed model. We characterize specialized regularization and network parameters with colored dotted lines and boxes with blue color, respectively.}\label{fig:overview}
\end{figure*}
\indent The second mechanism post-processes trained word embeddings to debias them after the training. An example of such post-processings is a linear projection of gender-neutral words toward a subspace, which is orthogonal to the gender direction vector defined by a set of gender-definition words \cite{bolukbasi16}. Another way of constructing the gender direction vector is using common names, e.g. \textit{john}, \textit{mary}, etc \cite{dev19}, while the previous approach used gender pronouns, such as \textit{he} and \textit{she}. In addition to the linear projections, \citet{dev19} utilizes other alternatives, such as flipping and subtraction, to reduce the gender bias more effectively. Beyond simple projection methods, \citet{kaneko19} proposed a neural network based encoder-decoder framework to add a regularization on preserving the gender-related information in feminine and masculine words. 

\section{Methodology}
\noindent Our model introduces 1) the siamese network structure \cite{Bromley94,weston2012deep} with an adapted gradient reversal layer for latent disentanglement and 2) the counterfactual data augmentation with geometric regularization for gender debiasing. We process the gender word pairs through the siamese network with auxiliary classifiers to reflect the inference of gender latent dimensions. Afterwards, we debias the gender-neutral words by locating it to be at the middle between a reconstructed pair of original gender latent variable and counterfactually generated gender latent variable. \\
\indent Same as previous researches \cite{kaneko19}, we divide a whole set of vocabulary $V$ into three mutually exclusive categories : \emph{feminine} word set $V_{f}$; \emph{masculine} word set $V_{m}$; and \emph{gender neutral} word set $V_{n}$, such that $V = V_{f} \cup V_{m} \cup V_{n}$ . In most cases, words in $V_{f}$ and $V_{m}$ exist in pairs, so we denote $\Omega$ as the set of feminine and masculine word pairs, such that $(w_{f}, w_{m}) \in \Omega$.

\subsection{Overall Model Structure}
\noindent Figure \ref{fig:overview} illustrates the overall structure of our proposed method for pre-trained word embeddings, which we named \textit{Counterfactual}-Debiasing, or \textit{CF}-Debias.
Eq. \eqref{eq:total_loss} specifies the entire loss function of the whole network parameters in Figure \ref{fig:overview}. The entire loss function is divided into two types of losses: $L_{ld}$ to be a loss for disentanglement and $L_{cf}$ to be a loss for counterfactual generation. $\lambda$ can be seen as a balancing hyper-parameter between two-loss terms.
\begin{align}
\label{eq:total_loss}
L = \lambda L_{ld}+ (1-\lambda) L_{cf},   0 \leq \lambda \leq 1
\end{align}
\indent Here, we use pre-trained word embeddings $\{\boldsymbol{w}_{i}\}_{i=1}^{V} \in \mathbb{R}^{d}$ for the debiasing mechanism. In the encoder-decoder framework, we denote the latent variable of $\boldsymbol{w}_{i}$ to be $\boldsymbol{z}_{i} \in \mathbb{R}^{l}$, which is mapped to the latent space by the encoding function, $E:\boldsymbol{w}_{i} \rightarrow \boldsymbol{z}_{i}$; and the decoding function, $D:\boldsymbol{z}_{i} \rightarrow \boldsymbol{\hat{w}}_{i}$. After the disentanglement of the latent space, $\boldsymbol{z}_{i}$ is divided into two parts, such that $\boldsymbol{z}_{i}$= $[\boldsymbol{z}^{s}_{i}, \boldsymbol{z}^{g}_{i}]$ : $\boldsymbol{z}^{s}_{i} \in \mathbb{R}^{l-k}$ is the semantic latent variable of $\boldsymbol{w}_{i}$; and $\boldsymbol{z}^{g}_{i} \in \mathbb{R}^{k}$ is the gender latent variable of $\boldsymbol{w}_{i}$, where $k$ is the pre-defined value for the gender latent dimension.\footnote{For the simplicity in notations, we skip the word-index $i$ in the losses of our proposed method.}

\subsection{Siamese Auto-Encoder for Latent Disentanglement}
\noindent This section provides the construction details of $L_{ld}$. Eq. \eqref{eq:total_dis_loss} defines the objective function for latent disentanglement as a linearly-weighted sum of the losses.
\begin{align}
\label{eq:total_dis_loss}
L_{ld} = \lambda_{se}L_{se} + \lambda_{ge}L_{ge} + \lambda_{di}L_{di} + \lambda_{re}L_{re}
\end{align}
\indent For the disentanglement, our fundamental assumption is maintaining the identical semantic information in $\boldsymbol{z}^{s}$ for the gender word pairs, $(w_{f}, w_{m}) \in \Omega$. Under this assumption, we introduce a latent disentangling method by utilizing the siamese auto-encoder with gender word pairs. The data structure of the gender word pairs provide an opportunity to adapt the siamese auto-encoder structure because the gender word pairs almost always have two words in pair\footnote{This structure can be expanded as our gender coverage changes.}. \\
\noindent \textbf{Semantic Latent Formulation} First, we regularize a pair of semantic latent variables $(\boldsymbol{z}^{s}_{f}, \boldsymbol{z}^{s}_{m})$, from a gender word pair, $(w_{f},w_{m})$, to be same by minimizing the squared $\ell_{2}$ distance as Eq. \eqref{eq:sim_loss}, since the semantic information of a gender word pair should be the same regardless of the gender.
\begin{align}
\label{eq:sim_loss}
L_{se} = \sum_{(w_{f}, w_{m}) \in \Omega} {\|\boldsymbol{z}_{m}^{s}-\boldsymbol{z}_{f}^{s}\|}^{2}_{2}
\end{align}
\textbf{Gender Latent Formulation} To formulate the gender-dependent latent dimensions, we introduce an auxiliary gender classifier, $C_r:\boldsymbol{z}^{g} \rightarrow [0,1]$, given in Eq. \eqref{eq:reg_loss}, and $C_r$ is asked to produce one in masculine words, labeled as $g_m=1$, and to produce zero in feminine words, $g_f=0$, respectively. After training, the output of $C_r$ can be an indicator of the gender information for each word.\footnote{We report the test performances of the gender classifier for gender-definition words, i.e., he, she, etc.; and gender-stereotypical words, i.e., doctor, nurse, etc., in Appendix D.}
\begin{align}
\label{eq:reg_loss}
L_{ge} & = -\sum_{w_{m} \in V_{m}} g_m\log C_{r}(\boldsymbol{z}_{m}^{g}) \nonumber \\&- \sum_{w_{f} \in V_{f}} (1-g_f)\log(1- C_{r}(\boldsymbol{z}_{f}^{g}))
\end{align}
\textbf{Disentanglement of Semantic and Gender Latent} The above two regularization terms do not guarantee the independence between the semantic and the gender latent dimensions. To enforce the independence between two latent dimensions, we introduce a \textit{Generator} with \textit{Gradient Reversal Layer} (GRL), $C_a:\boldsymbol{z}^{s} \rightarrow \boldsymbol{z}^{g}$ \cite{Ganin16}, which generates the gender latent dimension with the semantic latent dimension. 
\begin{figure}[]
	\centering
	\includegraphics[width=1.0\linewidth, height=1.55in]{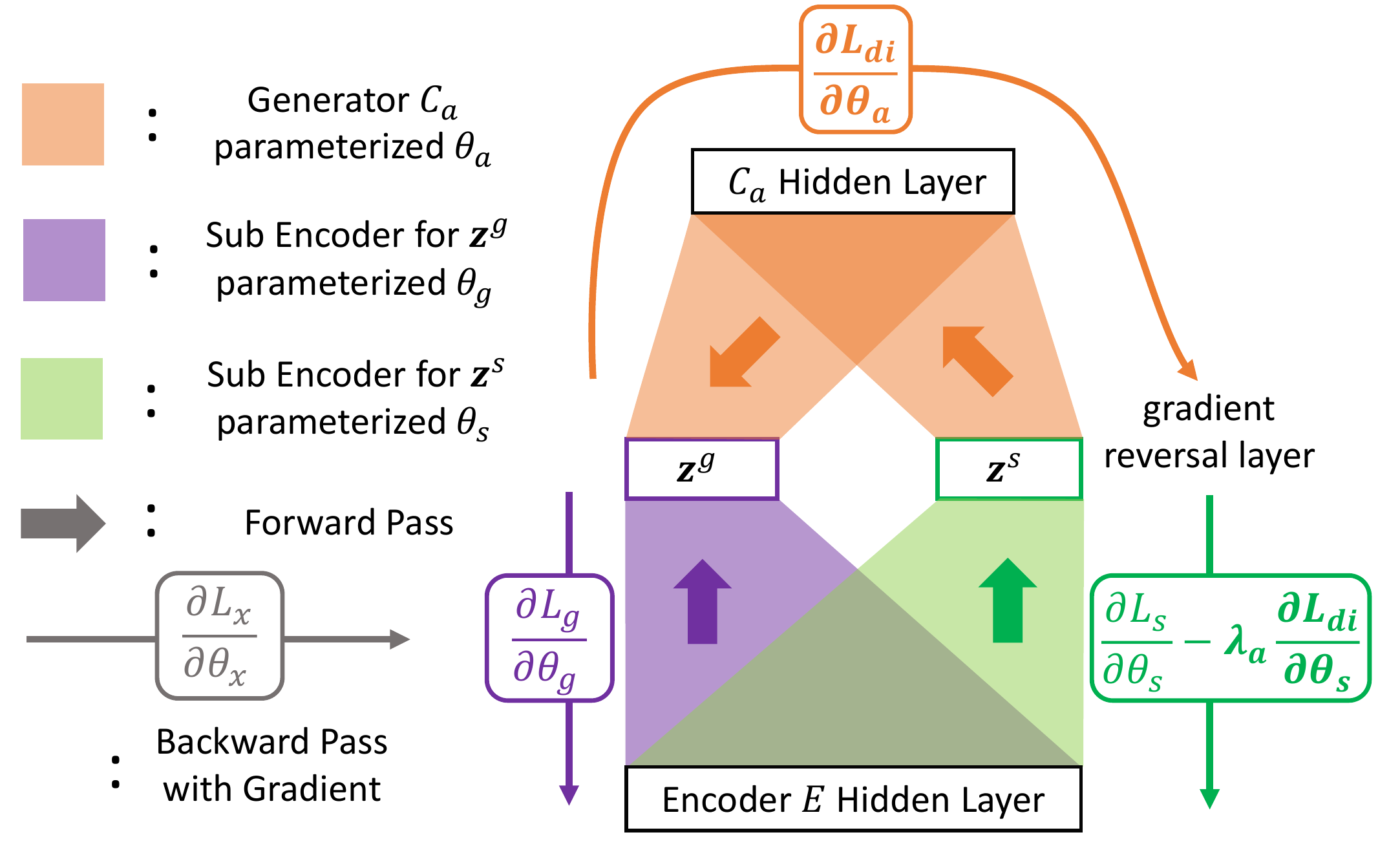}
	\caption{Gradient reversal layer utilized for the latent disentanglement. We follow similar description in \citet{Ganin16}}\label{fig:grl}
\end{figure}
We modify the flipping gradient idea of \cite{Ganin16} to the latent disentanglement between the semantic and the gender latent dimensions. The sufficient generation of $\boldsymbol{z}^g$ from $\boldsymbol{z}^s$ means that $\boldsymbol{z}^s$ has enough information on $\boldsymbol{z}^g$, so the generation should be prohibited to make $\boldsymbol{z}^g$ and $\boldsymbol{z}^s$ independent. 
Hence, our feedback of the gradient reversal layer is maximizing the loss of generating $\boldsymbol{z}^g$ from $\boldsymbol{z}^s$, which is represented as $L_{di}$ in Eq. \eqref{eq:adv_loss}. 
\begin{align}
\label{eq:adv_loss}
L_{di} = \sum_{w \in V} {\|C_a(\boldsymbol{z}^{s}) - \boldsymbol{z}^{g}\|}^{2}_{2}
\end{align}
\indent In the learning stage, the gradient of the encoder for $\boldsymbol{z}^s$, which is parameterized as $\theta_{s}$, becomes the summation of 1) $\frac{\partial L_{s}}{\partial \theta_{s}}$, which is the gradient for the loss $L_{s}$, the latent disentanglement losses of the encoder for $\boldsymbol{z}^s$ excluding $L_{di}$
; and 2)  $-\lambda_{a} \frac{\partial L_{di}}{\partial \theta_{s}}$, which is the $\lambda_{a}$-weighted negative gradient of the loss $L_{di}$ which is reversed after passing the GRL, because we intend to train the encoder for $\boldsymbol{z}^s$ by preventing the generation of $\boldsymbol{z}^g$. Eq. \eqref{eq:adv_loss} specifies the loss function for the disentanglement by GRL, and Eq. \eqref{eq:der_adv_loss} specifies the reversed gradient, see Figure \ref{fig:grl}.
\begin{align}
\label{eq:der_adv_loss}
\frac{\partial L_{ld}}{\partial \theta_{s}} = \frac{\partial L_{s}}{\partial \theta_{s}}-\lambda_{a} \frac{\partial L_{di}}{\partial \theta_{s}}
\end{align}
\textbf{Reconstruction} We add the reconstruction loss given in Eq. \eqref{eq:rec_loss} for this encoder-decoder framework.
\begin{align}
\label{eq:rec_loss}
L_{re} = \sum_{w \in V} {\|\boldsymbol{w} - \hat{\boldsymbol{w}}\|}^{2}_{2}
\end{align}
\subsection{Gender-Counterfactual Generation}
\noindent This section provides the construction details of $L_{cf}$. Same as $L_{ld}$, We define the objective function for the counterfactual generation as the linearly-weighted sum of the losses, introduced in this section, as in Eq. \eqref{eq:total_cnt_loss}. 
\begin{align}
\label{eq:total_cnt_loss}
L_{cf} = \lambda_{mo}L_{mo} + \lambda_{mi}L_{mi}
\end{align}
\indent Unlike the gender word pairs, a word in the gender neutral word set $w_{n} \in V_{n}$ utilizes a counterfactual generator, $C_{g}:\boldsymbol{z}^{g}_{n} \rightarrow \neg{\boldsymbol{z}^{g}_{n}}$, which converts the original gender latent, $\boldsymbol{z}^{g}_{n}$, to the opposite gender, $\neg{\boldsymbol{z}^{g}_{n}}$. It should be noted that $C_{g}$ is only activated for optimizing the losses in $L_{cf}$, which assumes that other parameters learned for the latent disentanglement are freezed. \\
\indent To switch $\boldsymbol{z}^{g}_{n}$, we utilize a prediction from the gender classifier, $C_{r}$, which is trained through the disentanglement loss. The modification loss, $L_{mo}$, originates from indicating the opposite gender with $\boldsymbol{z}^{g}_{n}$ by $C_{r}$, see Eq. \eqref{eq:cnt_loss}. For instance, if $C_{r}$ returns 0.8 for the original gender latent, $\boldsymbol{z}^{g}_{n}$, then we regularize the virtually generated gender latent, $\neg{\boldsymbol{z}^{g}_{n}}$, to lead $C_{r}$ to return 0.2. 
\begin{align}
\label{eq:cnt_loss}
L_{mo} = \sum_{w_{n} \in V_{n}} {\|C_{r}\left(\neg{\boldsymbol{z}^{g}_{n}}\right)-(1-C_{r}(\boldsymbol{z}^{g}_{n}))\|}^{2}_{2}
\end{align}

\indent While Eq. \eqref{eq:cnt_loss} focuses on the gender latent switch, Eq. \eqref{eq:self_loss} emphasizes the minimal change of the gender latent, $\boldsymbol{z}^{g}_{n}$. The combination of these two losses guides to the switched gender latent variable that is close to the original gender latent variable for regularizing the counterfactural generation.
\begin{align}
\label{eq:self_loss}
L_{mi} = \sum_{w_{n} \in V_{n}} {\|\neg{\boldsymbol{z}^{g}_{n}}-\boldsymbol{z}^{g}_{n}\|}^{2}_{2}
\end{align}


Though we keep the semantic latent variable, $\boldsymbol{z}^s$, and switch the gender latent variable, $\boldsymbol{z}^g$, to generate the gender-counterfactual word embedding, their concatenation during decoding can be vulnerable to the semantic information changes because of variances in the individual latent variables. 
Consequently, we constrain that the reconstructed word embedding with the counterfactual gender latent, $\hat{\boldsymbol{w}}_{cf}$, differs only in the gender information from $\hat{\boldsymbol{w}}_{n}$, which is the reconstructed word embedding with the original gender latent. \\
\textbf{Linear Alignment} For this purpose, we introduce the linear alignment, which regularizes $\hat{\boldsymbol{w}}_{n}-\hat{\boldsymbol{w}}_{cf}$ by measuring the alignment to the gender direction vector $\boldsymbol{v}_{g}$ in Eq. \eqref{eq:gender_direction}, which is an averaged gender difference vector from the gender word pairs.
\begin{align}
\label{eq:gender_direction}
\boldsymbol{v}_{g} = \frac{1}{|\Omega|}\sum_{(w^{i}_{f}, w^{i}_{m}) \in \Omega} \left(\hat{\boldsymbol{w}}^{i}_{m} - \hat{\boldsymbol{w}}^{i}_{f}\right)
\end{align}
This regularization suggests that we constrain the embedding shift of the gender-neutral word to be the direction of $\boldsymbol{v}_{g}$. This alignment can be accomplished by maximizing the absolute inner product between $\hat{\boldsymbol{w}}_{n}-\hat{\boldsymbol{w}}_{cf}$ and $\boldsymbol{v}_{g}$ as given in Eq. \eqref{eq:align_loss}. We introduce $CF$-Debias-LA, which adds the below linear alignment regularization, $\lambda_{la}L_{la}$, to $L_{cf}$.
\begin{align}
\label{eq:align_loss}
L_{la} = & \sum_{w_{n} \in V_{n}} -|\boldsymbol{v}_{g} \boldsymbol{\cdot} (\hat{\boldsymbol{w}}_{n}-\hat{\boldsymbol{w}}_{cf})|
\end{align}
\textbf{Kernelized Alignment}
While the linear alignment computes the gender direction vector $\boldsymbol{v}_{g}$ as a simple average, the gender information of word embedding can have a nonlinear structure. Therefore, we introduce the \textit{kernelized alignment}, which enables the nonlinear alignment between 1) $\hat{\boldsymbol{w}}^{i}_{m} - \hat{\boldsymbol{w}}^{i}_{f}$ of each gender word pair $(w^{i}_{f}, w^{i}_{m})$ and 2) $\hat{\boldsymbol{w}}_{n}-\hat{\boldsymbol{w}}_{cf}$ of gender-neutral words $w_n$.

We hypothesize a nonlinear mapping function $f$, which projects a word embedding $\boldsymbol{w}_{i} \in \mathbb{R}^{d}$ into a newly introduced feature space, $f(\boldsymbol{w}_{i}) \in \mathbb{R}^{m}$. We can utilize the kernel trick \cite{1998nonlinear} for computing pairwise operation on the nonlinear space introduced by $f$ . Let ${\Ked}(\boldsymbol{w},\boldsymbol{w}') = f(\boldsymbol{w})\cdot f(\boldsymbol{w}')$ be a kernel representing an inner-product of two vectors in the feature space. Also, we set $\phi_{k}$ to be $k$-th eigenvector for the projected outputs of the given embeddings $\{f(\boldsymbol{w}_{i})\}_{i=1}^{N}$. By following Appendix A, ${PC}_{k}$ is the $k$-th principal component of new word embedding $\boldsymbol{w}'$ on the introduced feature space: ${PC}_k=f(\boldsymbol{w}')\cdot\phi_{k}$. 
Then, we find the $k$-th principal component for embedding $\boldsymbol{w}'$ as given in Eq. \eqref{eq:principal_component}, when $a^{i}_{k}$ is $i$-th component of $k$-th eigenvector of $\boldsymbol{K}$, which is a $N \times N$ kernel matrix of given data.
\begin{align}
\label{eq:principal_component}
{PC}_{k}=f(\boldsymbol{w}') \cdot \phi_{k} =  &\Sigma^{N}_{i=1}{a^{i}_{k}f(\boldsymbol{w}_{i}) \cdot f(\boldsymbol{w}')} \nonumber \\= &\Sigma^{N}_{i=1}{a^{i}_{k}\Ked(\boldsymbol{w}_{i},\boldsymbol{w}')}
\end{align}
Substituting the inner product in Eq. \eqref{eq:align_loss} with Eq. \eqref{eq:kernel_alignment}, we design the nonlinear alignment between the gender difference vector, $\hat{\boldsymbol{w}}_{m} - \hat{\boldsymbol{w}}_{f}$, and the gender neutral vector, $\hat{\boldsymbol{w}}_{n}-\hat{\boldsymbol{w}}_{cf}$, by maximizing the Top-$K$ kernel principal components as Eq. \eqref{eq:kernel_alignment}. We introduce $CF$-Debias-KA, which adds the kernelized alignment regularization, $\lambda_{ka}L_{ka}$, to $L_{cf}$. We use Radial Basis Function kernel for our experiment.
\begin{align}
\label{eq:kernel_alignment}
& L_{ka} = \nonumber- \sum^{K}_{k=1}\sum_{w_{n} \in V_{n}}\sum_{(w^{i}_{f}, w^{i}_{m}) \in \Omega} \\ & a^{i}_{k}\Ked\left(\hat{\boldsymbol{w}}^{i}_{m} - \hat{\boldsymbol{w}}^{i}_{f},\hat{\boldsymbol{w}}_{n}-\hat{\boldsymbol{w}}_{cf} \right)
\end{align}


\subsection{Post-Processing by the Word's Category}
\noindent After learning the network parameters, we post-process words by its categories of $V_{f}$, $V_{m}$, and $V_{n}$. We gender-neutralize the embedding vector of $w_{n} \in V_{n}$ by relocating the vector to the middle point of the reconstructed original-counterfactual pair embeddings, such that $\boldsymbol{w} :=  \frac{\hat{\boldsymbol{w}}_{cf}+\hat{\boldsymbol{w}}_{n} }{2}= \hat{\boldsymbol{w}}_{neu}$. We utilize a reconstructed word embedding which preserves the gender information in embedding space, $\boldsymbol{w} :=  \hat{\boldsymbol{w}}_{f}$ for $w_{f} \in V_{f}$ and $\boldsymbol{w} :=  \hat{\boldsymbol{w}}_{m}$ for  $w_{m} \in V_{m}$. For each $w \in V_{f}\cup V_{m}$, we can safely preserve gender information of given word by using reconstructed embedding such that $\boldsymbol{w} :=  \hat{\boldsymbol{w}}$.


\begin{table*}[]
\begin{adjustbox}{width=\textwidth,height=0.85in}
\begin{tabular}{lccc ccc ccc ccc ccc ccc}
\toprule
& \multicolumn{6}{c}{\textbf{English (GloVe)}} & \multicolumn{6}{c}{\textbf{Spanish (Fasttext)}} &\multicolumn{6}{c}{\textbf{Korean (Fasttext)}} \\
\cmidrule(lr){2-7} \cmidrule(lr){8-13} \cmidrule(lr){14-19} 
&\multicolumn{3}{c}{\textsf{Sembias}} & \multicolumn{3}{c}{\textsf{Sembias subset}} &\multicolumn{3}{c}{\textsf{Sembias}} & \multicolumn{3}{c}{\textsf{Sembias subset}} &\multicolumn{3}{c}{\textsf{Sembias}} & \multicolumn{3}{c}{\textsf{Sembias subset}} \\
\cmidrule(lr){2-4}  \cmidrule(lr){5-7} \cmidrule(lr){8-10}  \cmidrule(lr){11-13} \cmidrule(lr){14-16}  \cmidrule(lr){17-19} 
Embeddings &Def $\uparrow$ & Stereo$\downarrow$ & None $\downarrow$ & Def $\uparrow$ & Stereo $\downarrow$ & None $\downarrow$ &Def $\uparrow$ & Stereo$\downarrow$ & None $\downarrow$ & Def $\uparrow$ & Stereo $\downarrow$ & None $\downarrow$ &Def $\uparrow$ & Stereo $\downarrow$ & None $\downarrow$ & Def $\uparrow$ & Stereo $\downarrow$ & None $\downarrow$ \\
\midrule
Original      & 80.22       & 10.91       & 8.86 & 57.5       & 20.0       & 22.5  
&  $70.98^{\dagger}$ & $17.38^{\dagger}$ & $11.63^{\dagger}$ & $84.61^{\dagger}$ & $11.86^{\dagger}$ & $3.52^{\dagger}$& 
$80.38^{\dagger}$ & $7.48^{\dagger}$  & $12.14^{\dagger}$ & 76.26 & 8.87  & 14.88 \\
Hard-Debias &  $87.95^{\ast}$       &   8.41       &  $3.64^{\ast}$ &  50.0       &   32.5       &  17.5 
& 41.76 & 27.55 & 30.68 & 21.12  & 38.54 & 40.33&
41.39 & 15.31 & 43.30 & $\mathbf{89.23}^{\ast}$ & $2.62^{\ast}$  & $\mathbf{8.15}^{\ast}$  \\
GN-Debias   &   $97.73^{\dagger\ast}$      &    $1.36^{\dagger \ast}$        &  $0.91^{\dagger \ast}$  &   $75.0^{\dagger}$      &    15.0        &  10.0  
& ---- & ---- & ---- & ---- & ---- & ---- & ---- & ---- & ---- & ---- & ---- & ----\\
ATT-Debias   &   80.22      &    10.68        &  9.09 &   60.0      &    17.5        &  22.5   
&$75.23^{\ast\dagger}$ & $13.02^{\ast\dagger}$ & $11.74^{\dagger}$ & $83.44^{\dagger}$  & $9.80^{\dagger\ast}$   & $6.76^{\dagger}$  & 
$82.98^{\dagger\ast}$ & $7.70^{\dagger}$  & $\mathbf{9.33}^{\dagger\ast}$  & $79.59^{\ast}$ & 8.87  & $11.55^{\ast}$ \\ 
CPT-Debias   &   73.63      &    5.68        &  20.68 &   45.0      &    12.5        &  42.5   
& $69.62^{\dagger}$ & $18.26^{\dagger}$ & $12.11^{\dagger}$ & $84.62^{\dagger}$  & $11.86^{\dagger}$ & $3.52^{\dagger}$  & 
$61.31^{\dagger}$ & $10.57^{\dagger}$ & $28.12^{\dagger}$ & 38.52 & 15.76 & 45.72 \\ 
AE-Debias   &   84.09      &    7.95         &  7.95  &   $65.0^{\dagger}$      &    15.0        &  20.0  
&$73.19^{\ast\dagger}$ & $15.56^{\ast\dagger}$ & $11.26^{\dagger}$ & $86.38^{\dagger\ast}$  & $10.10^{\dagger\ast}$  & $3.52^{\dagger}$  &  
$57.66^{\dagger}$ & $11.91^{\dagger}$ & $30.44^{\dagger}$ & 55.72 & 10.76 & 33.53 \\
AE-GN-Debias   &   $98.18^{\dagger\ast}$      &    $1.14^{\dagger \ast}$        &  $0.68^{\dagger \ast}$ &   $80.0^{\dagger\ast}$      &    $12.5^{\dagger}$        &  7.5  & ---- & ---- & ---- & ---- & ---- & ---- & ---- & ---- & ---- & ---- & ---- & ---- \\
GP-Debias  &   84.09      &    8.18        &  7.73 &   $65.0^{\dagger}$      &    15.0       &  20.0   
& $72.93^{\dagger\ast}$ & $15.87^{\dagger\ast}$ & $\mathbf{11.19}^{\dagger\ast}$ & $86.37^{\dagger\ast}$  & $10.09^{\dagger\ast}$ & $3.52^{\dagger}$ &
$55.85^{\dagger}$ & 15.62 & $28.53^{\dagger}$ & 68.00 & 16.19 & 15.81 \\
GP-GN-Debias   &   $98.41^{\dagger\ast}$      &    $1.14^{\dagger \ast}$        &  $0.45^{\dagger \ast}$  &   $82.5^{\dagger\ast}$      &    $12.5^{\dagger}$        &  $5.0^{\ast}$ & ---- & ---- & ---- & ---- & ---- & ---- & ---- & ---- & ---- & ---- & ---- & ---- \\
\midrule
CF-Debias   & $98.18^{\dagger\ast}$ & $0.68^{\dagger\ast}$ & $1.13^{\dagger\ast}$ & $80.0^{\dagger\ast}$  & $7.5{\dagger}$  & 12.5  
&  $78.93^{\dagger\ast}$ & $\mathbf{3.83}^{\dagger\ast}$  & $17.23^{\dagger}$ & $96.15^{\dagger\ast}$  & $\mathbf{0.0}^{\dagger\ast}$    & $3.85^{\dagger}$  & $83.02^{\dagger\ast}$ & $2.44^{\dagger\ast}$   &  $14.53^{\dagger}$ &  $80.98^{\ast}$ &  $\mathbf{0.0}^{\dagger \ast}$ & 19.02 \\
 CF-Debias-LA  &   $\mathbf{100.00}^{\dagger\ast}$      &    $\mathbf{0.00}^{\dagger \ast}$        &  $\mathbf{0.00}^{\dagger \ast}$ &   $\mathbf{100.0}^{\dagger\ast}$      &    $\mathbf{0.0}^{\dagger\ast}$        &  $\mathbf{0.0}^{\dagger\ast}$  
& $69.33^{\dagger}$ & $9.05^{\dagger\ast}$  & $21.61^{\dagger}$ & $\mathbf{100.0}^{\dagger\ast}$    & $\mathbf{0.0}^{\dagger\ast}$    & $\mathbf{0.0}^{\dagger\ast}$    & $\mathbf{85.07}^{\dagger\ast}$ & $2.37^{\dagger\ast}$   &  $12.5^{\dagger}$ &  $88.04^{\ast}$ &  $\mathbf{0.0}^{\dagger \ast}$ & $11.95^{\ast}$  \\
 CF-Debias-KA  & $92.04^{\dagger\ast}$  & $3.41^{\dagger\ast}$   &  $4.55^{\ast}$ & 62.5 & 17.5 & 20.0   
&$\mathbf{80.35}^{\ast\dagger}$ & $6.73^{\ast\dagger}$  & $12.91^{\dagger}$ & $\mathbf{100.0}^{\dagger\ast}$   & $\mathbf{0.0}^{\dagger\ast}$     & $\mathbf{0.0}^{\dagger\ast}$  & $84.28^{\dagger\ast}$ & $\mathbf{2.09}^{\dagger\ast}$   &  $13.62^{\dagger}$ &  $82.27^{\ast}$ &  $2.38^{\ast}$ & 15.35   \\
\bottomrule
\end{tabular}
\end{adjustbox}
\caption{Percentage of predictions of each category on sembias analogy task, for each language. ${\dagger}$ and ${\ast}$ denote the statistically significant differences for Hard-Debias and Original embedding, respectively. The best model is indicated as boldface. We denote "---" for the skipped cases, whose methods are closely tied to GloVe embedding.}
\label{table:sembias_1}
\end{table*}

\section{Experiments}
\subsection{Datasets and Experimental Settings}
\noindent We used the set of gender word pairs created by \citet{zhao18} as $V_{f}$ and $V_{m}$, respectively. 
All models utilize GloVe on 2017 January dump of English Wikipedia with 300-dimension embeddings for 322,636 unique words. Additionally, to investigate the debiasing effect on languages other than English; we conducted one of the debiasing experiments for Spanish, which is the Subject-Verb-Object language as English; and Korean, one of the Subject-Object-Verb language. We used Fasttext \cite{fasttext} for experiments of Spanish and Korean. Accordingly, we excluded the baselines, whose methods are closely tied to GloVe, for the experiments of other languages. We specify the dimensions of $\boldsymbol{z}$, $l$, as 300, which is divided into 295 semantic latent dimensions and 5 gender latent dimensions. Also, we utilize the sequential hyper-parameter schedule, which updates the weight for $L_{ld}$ more at the initial step and gradually increases updating the weight for the $L_{cf}$, by changing $\lambda$ in Eq. \eqref{eq:total_loss} from 1 to 0. Further information on experimental settings can be found in Appendix G.
\subsection{Baselines}
\noindent We compare our proposed model with below baseline models, and we utilize the authors' implementations.\footnote{We provided link of the authors' implementations in Appendix H.}
Hard-Debias \cite{bolukbasi16} utilizes linear projection technique for gender-debiasing. GN-Debias \cite{zhao18} trains the word embedding from scratch by preserving the gender information into the specific dimension and regularizing the other dimensions to be gender-neutral. CPT-Debias \cite{conceptor2019} introduces a debiasing mechanism by utilizing the conceptor matrix. ATT-Debias \cite{dev19} defines gender subspace with common names and proposes the subtraction and the linear projection methods based on gender subspace.\footnote{We use the subtraction method as an ATT-Debias.} AE-Debias and AE-GN-Debias \cite{kaneko19} utilize the autoencoder structure for debiasing, and utilize the original word embedding and GN-Debias, respectively. Besides, GP-Debias and GP-GN-Debias adopt additional losses to neutralize gender bias and preserve gender information for gender-definition words.
\subsection{Quantitative Evaluation for Debiasing}
\subsubsection{Sembias Analogy Test}
\label{sec:sembias}
\noindent We perform the \textit{Sembias} gender analogy test \cite{zhao18, jurgens2012semeval} to evaluate the degree of gender bias in embeddings. The \textit{Sembias} dataset in English contains 440 instances, and each instance consists of four-word pairs : 1) a gender-definition word pair (Def), 2) a gender-stereotype word pair (Stereo), and 3,4) two none-type word pairs (None). We test models by calculating the linear alignment between each word pair difference vector, $\overrightarrow{a}-\overrightarrow{b}$; and $\overrightarrow{he}-\overrightarrow{she}$, which we refer to as \textit{Gender Direction}. This test regards an embedding model to be better debiased if the alignment is larger for the word pair of Def compared to the word pairs of None and Stereo. By following the past practices, we test models with 40 instances from a subset of \textit{Sembias}, whose gender word pairs are not used for training. To investigate the result of \textit{Sembias} analogy test in Spanish and Korean, we translated the words in \textit{Sembias} into the other languages with human corrections.\\
\indent Table \ref{table:sembias_1} shows the percentages of the largest alignment with \textit{Gender Direction} for all instances. For English, CF-Debias-LA selects all the pairs of Def, which shows the sufficient maintenance of the gender information for those words. Also, CF-Debias-LA selects neither stereotype nor none-type words, so the difference vectors of Stereo and None always have less alignment to \textit{Gender Direction} than the difference vectors of Def. We further refer to the experimental settings of Spanish and Korean in Appendix J.

\begin{table*}[ht!]
\begin{center}
\begin{adjustbox}{width=0.85\textwidth,height=0.94in}
	\begin{tabular}{l cc cc cc cc cc}
	\toprule
	 & \multicolumn{2}{c}{\textsf{career vs family}} & \multicolumn{2}{c}{\textsf{math vs art}} & \multicolumn{2}{c}{\textsf{science vs art}} & \multicolumn{2}{c}{\textsf{intellect vs appear}} & \multicolumn{2}{c}{\textsf{strong vs weak}}\\
	\cmidrule(lr){2-3}  \cmidrule(lr){4-5} \cmidrule(lr){6-7} \cmidrule(lr){8-9} \cmidrule(lr){10-11}
	Embeddings&\large p-value &\large $d$   &\large p-value &\large $d$   &\large p-value &\large $d$  &\large p-value &\large $d$   &\large p-value &\large $d$ \\
	\midrule
	Original & 0.000          & 1.605          & 0.276          & 0.494           & 0.014          & 1.260          & 0.009          & 0.706           & 0.067          & 0.640          \\
	Hard-Debias &0.100 & 0.842 & 0.090          & -1.043          & 0.003          & -0.747         & \underline{0.693} & \underline{-0.121} & 0.255 & 0.400 \\
	GN-Debias & 0.000          & 1.635          & 0.726 & -0.169 & 0.081          & 1.007          & 0.037          & 0.595           & 0.083          & 0.620          \\
	ATT-Debias & \underline{0.612}              & \underline{0.255}              & 0.007              & -0.519               & 0.000              & 0.843              & 0.129              & 0.440               & 0.211              & 0.455              \\
	CPT-Debias & 0.004           & 1.334         & 0.058          & 1.029            & 0.000              & 1.417         & 0.001              & 0.906       & \underline{0.654}      & \underline{-0.172}              \\
	AE-Debias & 0.000          & 1.569          & 0.019          & 0.967           & 0.024          & 1.267          & 0.007          & 0.729           & 0.027          & 0.763          \\
	AE-GN-Debias &0.001          & 1.581          & 0.716          & 0.317           & 0.139          & 0.639          & 0.006          & 0.770           & 0.028          & 0.585          \\
	GP-Debias & 0.000          & 1.567          & 0.019          & 0.966           & 0.027          & 1.253          & 0.006          & 0.733           & 0.028          & 0.758          \\
	GP-GN-Debias & 0.000          & 1.599          & \textbf{0.932} & \underline{0.109}  & 0.251 & 0.591 & 0.004          & 0.791           & 0.098          & 0.610          \\
	\midrule
	CF-Debias& 0.210   & 0.653  & 0.759 & 0.261 & $\underline{0.725}$  & $\underline{-0.363}$ & 0.256  &-0.328 & 0.305 &0.371  \\
	 CF-Debias-LA  & \textbf{0.874} & \textbf{-0.089} & 0.669          & -0.125          & 0.360 & 0.480 & 0.678 & -0.124  & \textbf{0.970} & \textbf{0.013} \\
	 CF-Debias-KA  & 0.196 & 0.673 &   $\underline{0.887}$       &  $\textbf{0.083}$    & $\textbf{0.919}$ & $\textbf{-0.235}$  & \textbf{0.893}  & \textbf{-0.039} & 0.373 & 0.338 \\
\bottomrule
\end{tabular}
\label{table:weat1}
\end{adjustbox}
	
\caption{WEAT hypothesis test results for five gender-stereotypical word categories. The best and second-best models are indicated as boldface and underline, respectively. The absolute value of the effect size denotes the degree of bias. A value of $d$ closer to 0 means that there is no gender bias.}
\label{table:weat1}
\end{center}
\end{table*}

\subsubsection{WEAT}
\label{sec:weat}
\noindent We apply the Word Embedding Association Test (WEAT) \cite{weat2017} for debiasing test. WEAT uses permutation test to compute the effect size ($d$) and p-value in Table \ref{table:weat1}, as a measurement of the bias in word embeddings. The effect size computes differential association of the sets of stereotypical target words, i.e. \textit{career vs family}, and the gender word pair sets from \citet{chaloner2019}. A higher value of effect size indicates a higher gender bias between the two sets of target words. The p-value is used to check the significant level of bias. We provide the detailed description of WEAT in Appendix C. The variations of our method show the best performances for whole categories except \textit{math vs art}, see Table \ref{table:weat1}.
\begin{table}[h]
\begin{center}
	\begin{adjustbox}{height=0.70in}
	\begin{tabular}{l cc}
	\toprule
	Embeddings& $\textsf{no gender bias}$  & $\textsf{semantic validity}$ \\
	\midrule
	Original & 0.447$\pm$\small{0.179}  &  $\mathbf{0.875}$$\pm$\small{0.132}    \\
	Hard-Debias & 0.491$\pm$\small{0.142} & 0.652$\pm$\small{0.123}    \\
	ATT-Debias& 0.610$\pm$\small{0.136} & 0.761$\pm$\small{0.131}    \\
	CPT-Debias& 0.552$\pm$\small{0.128} & 0.827$\pm$\small{0.138}    \\
	GP-GN-Debias  & 0.328$\pm$\small{0.241} & 0.421$\pm$\small{0.149}    \\
	\midrule
	 CF-Debias-LA   & $\mathbf{0.644}$$\pm$\small{0.124} & 0.683$\pm$\small{0.152}    \\
	 CF-Debias-KA & 0.615$\pm$\small{0.107} & 0.744$\pm$\small{0.142}    \\
	\bottomrule
\end{tabular}
\label{table:human}
\end{adjustbox}
\caption{Human-based evaluation for the gender bias and semantics of generated analogy, with standard deviation. The best model is indicated as boldface.}
\label{table:human}
\end{center}
\end{table}
\subsubsection{Analogy Test with Human based Validation}
\noindent We conducted a human experiment on the analogy generated by the debiased embeddings to evaluate the debiasing efficacy of each embedding. each embeddings generate a word based on the question "$a$ is to $b$ as $c$ is to what?", when words $a, b$ are selected from the gender word pairs of \textit{Sembias} dataset; and $c$ is given as a gender stereotypical word, i.e. {homemaker, housekeeper}, from \citet{bolukbasi16}. The answer word from each question is generated by ${argmax}_{d \in V}(\overrightarrow{d}\boldsymbol{\cdot} (\overrightarrow{c}{-}\overrightarrow{a}{+}\overrightarrow{b}))$. 18 Human subjects were asked to evaluate the generated analogies from two perspectives; 1) existence of gender bias in the analogy, 2) semantic validity of the analogy.\footnote{We enumerate the embeddings utilized in an experiment and detailed description of the human experiment in Appendix I.} Table \ref{table:human} shows that our method indicates the least gender bias while competitively maintaining the semantic validity.



\subsection{Debiasing Qualitative Analysis}
\begin{figure}[hbt!]
\centering
\begin{subfigure}[t]{0.45\columnwidth}
	\centering
    	\includegraphics[width=1.4in,height=0.98in]{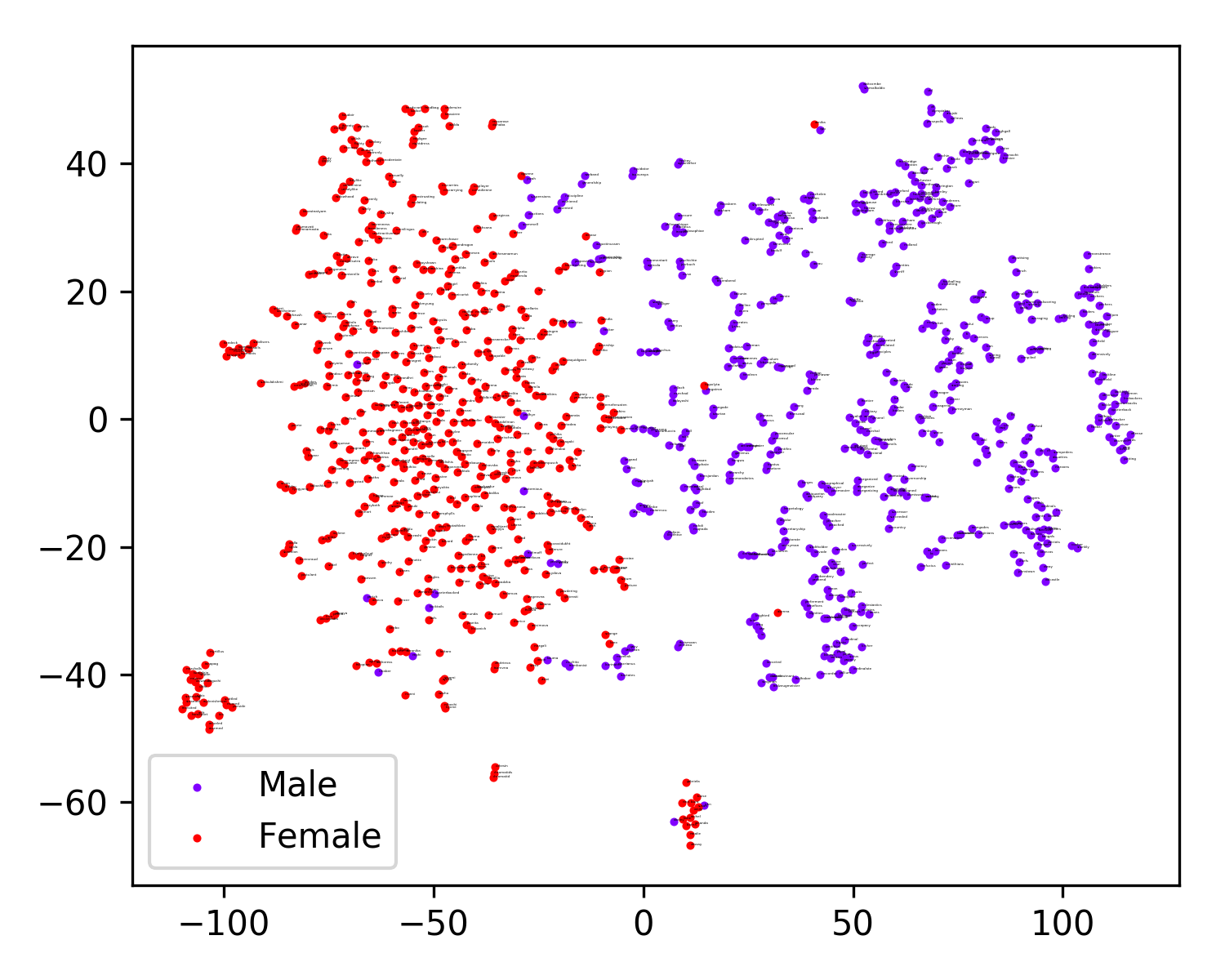}
	\captionsetup{labelformat=empty}
	\caption{\hspace{0.2cm} \small{Hard-Debias (89.4\%)}}

	\centering
    	\includegraphics[width=1.4in,height=0.99in]{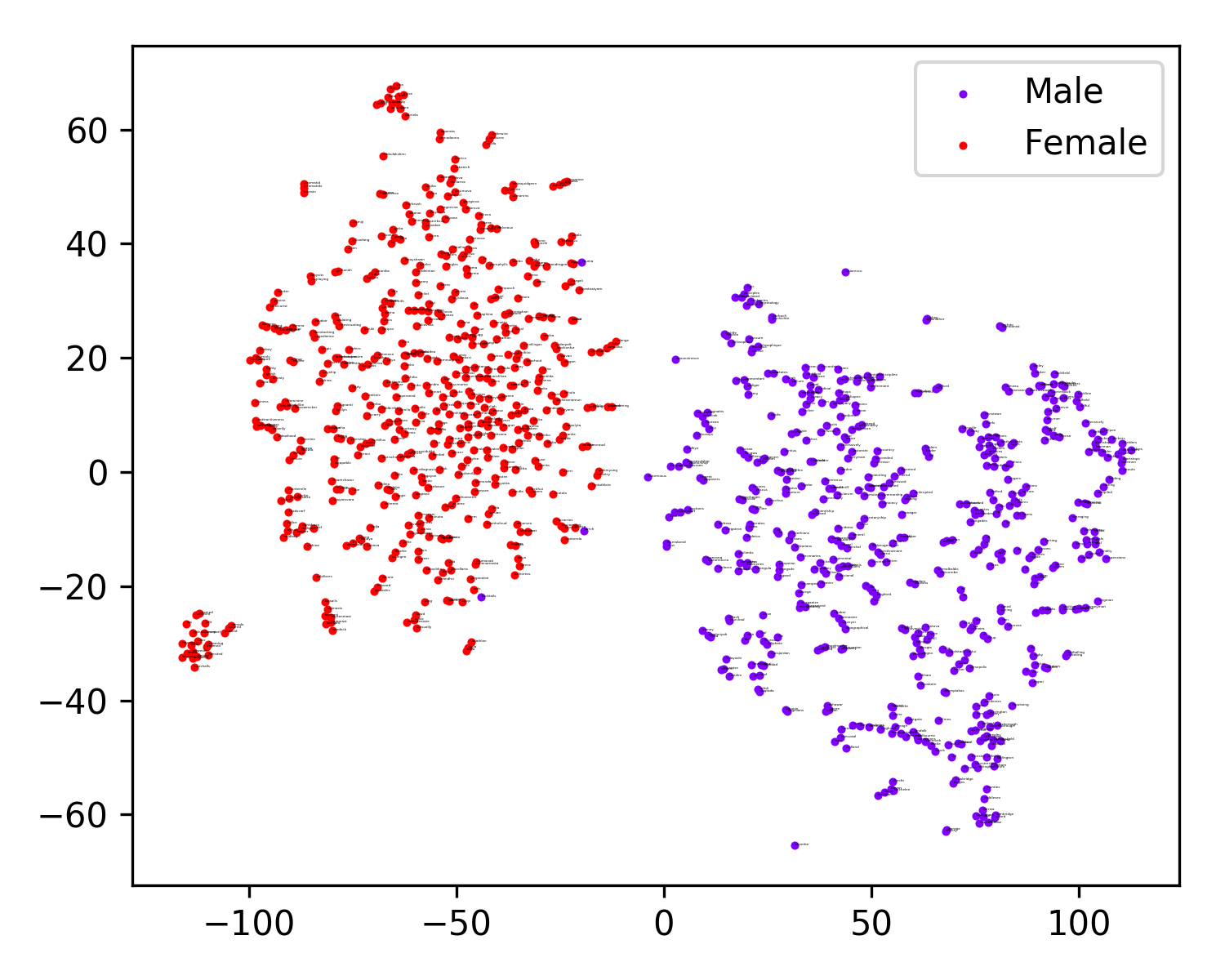}
	\captionsetup{labelformat=empty}
	\caption{\hspace{0.2cm} GP-GN-Debias (100.0\%)}

	\centering
	\includegraphics[width=1.4in,height=0.99in]{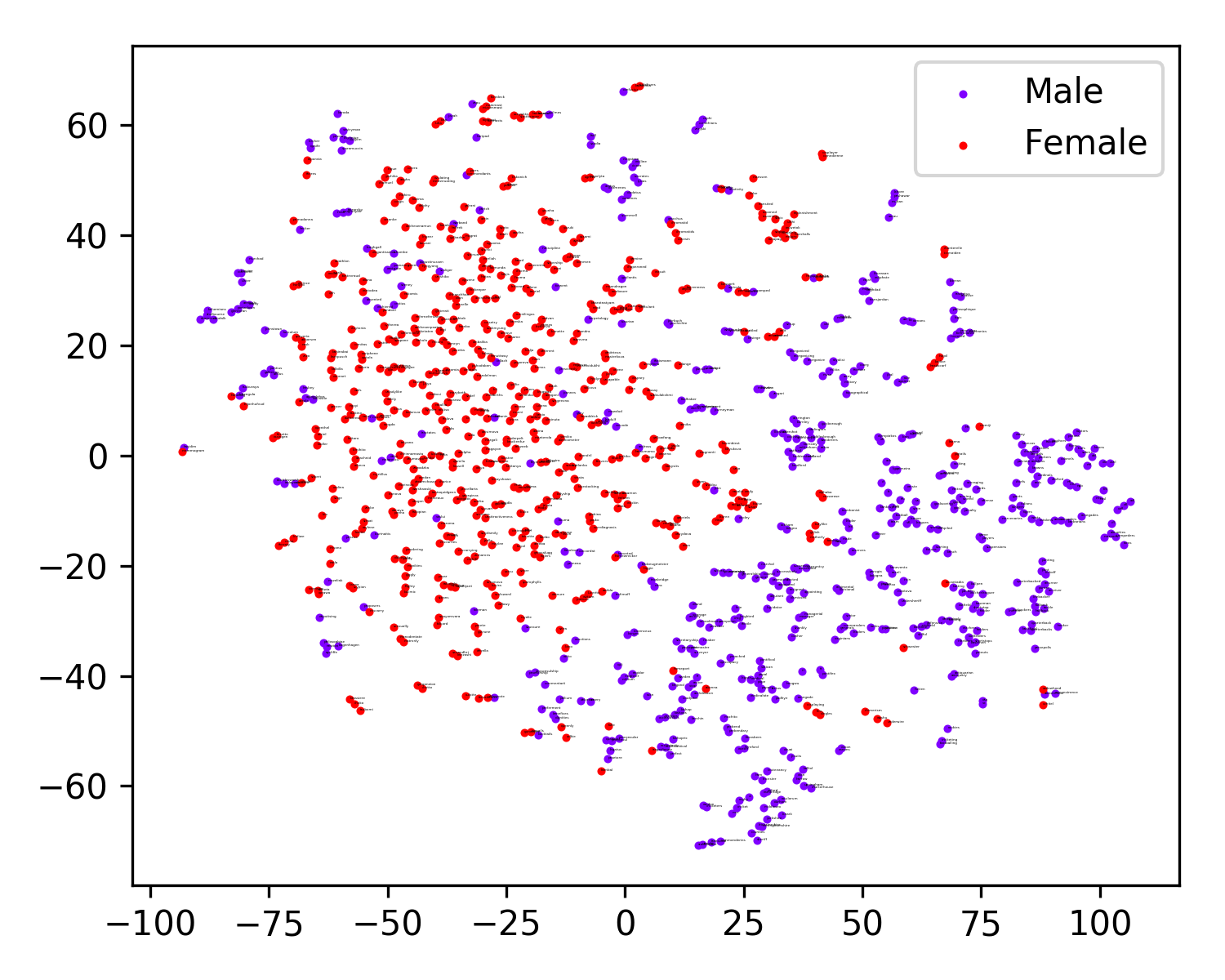}
	\captionsetup{labelformat=empty}
    	\caption{\hspace{0.2cm} CF-Debias-KA (76.8\%)}
\end{subfigure}\,
\centering
\begin{subfigure}[t]{0.51\columnwidth}
    \includegraphics[width=1.68in,height=0.99in]  
    {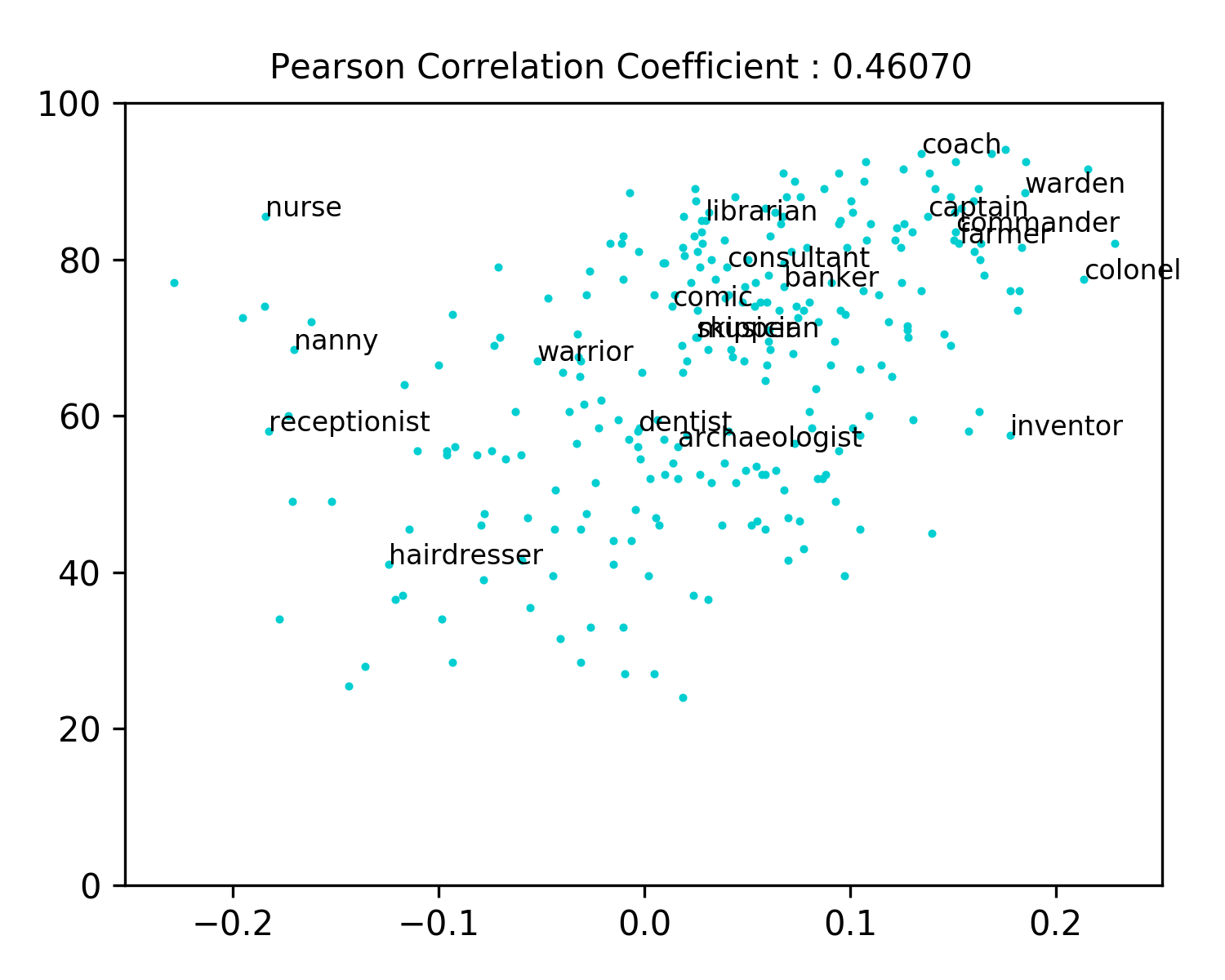}
	\captionsetup{labelformat=empty}
	\caption{ \hspace{0.3cm} \small{Hard-Debias (0.4607)}} 
    \includegraphics[width=1.68in,height=0.99in]
    {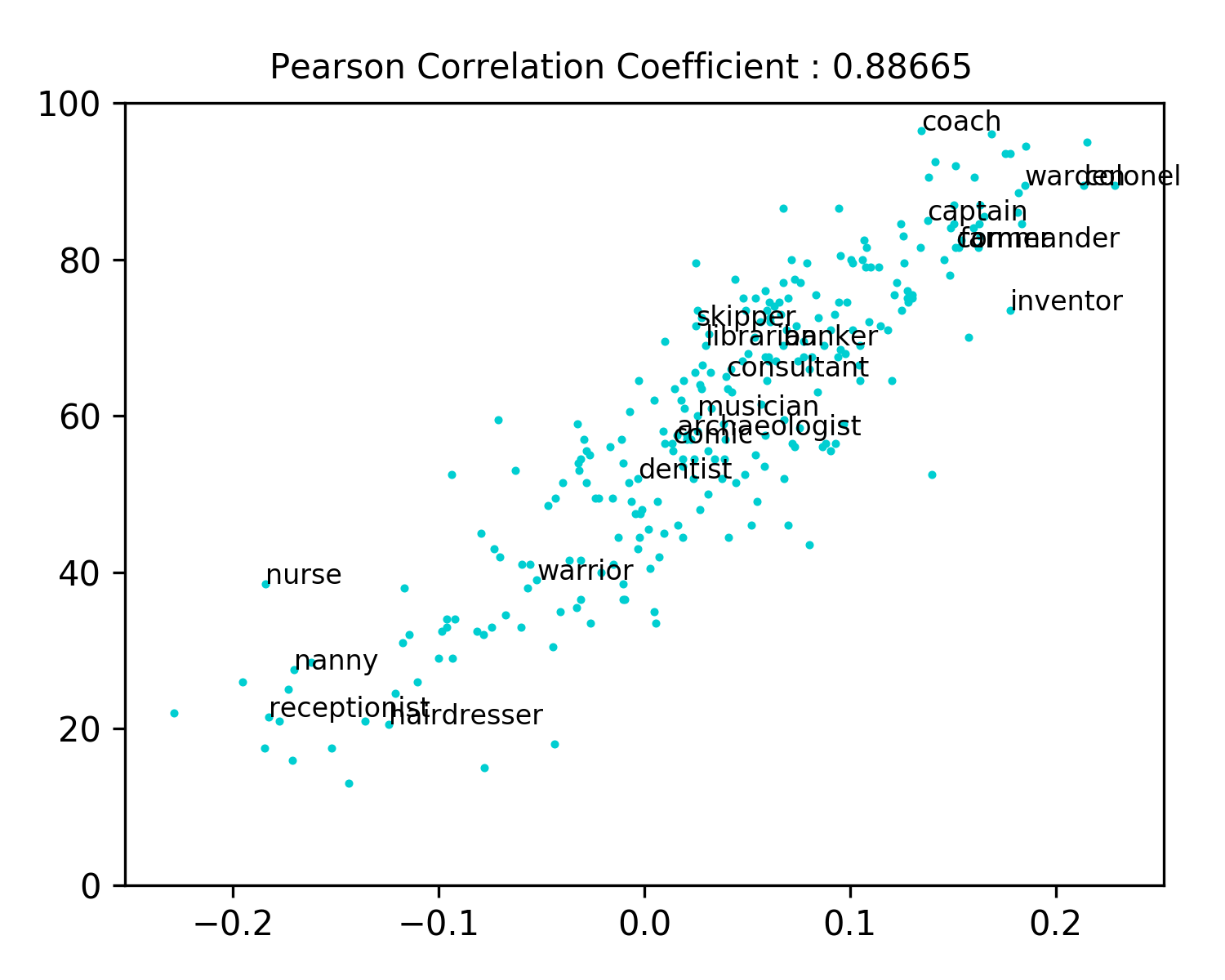}
	\captionsetup{labelformat=empty}
	\caption{ \hspace{0.3cm}GP-GN-Debias (0.8867)} 
    \includegraphics[width=1.68in,height=0.99in]
    {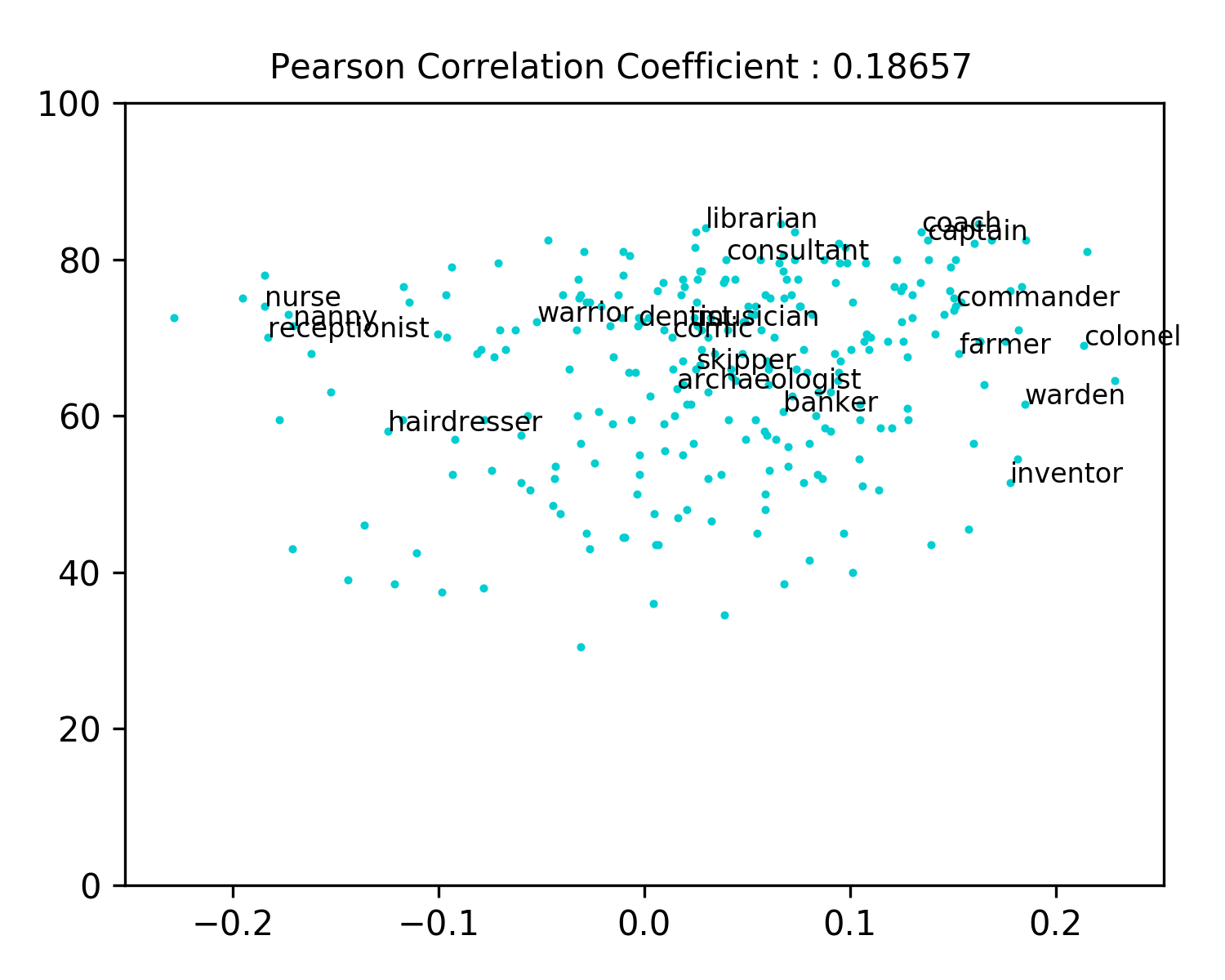}
	\captionsetup{labelformat=empty}
    	\caption{  \hspace{0.3cm} CF-Debias-KA (0.1866)} 
\end{subfigure}
\caption{The t-SNE views for 500 male, female-biased word embeddings from original embedding, with the cluster-based classification accuracy in parentheses. (left) The percentage of male neighbors for each profession as a function of original bias, with the Pearson correlation coefficient in parentheses. (right)}
\label{fig:nearest}
\end{figure}

\begin{table*}[ht!]
 \centering
 \begin{adjustbox}{width=0.85\textwidth}
 \begin{tabular}{l cc cc cc}
 \toprule
  & \multicolumn{2}{c}{\textsf{POS Tagging}} & \multicolumn{2}{c}{\textsf{POS Chunking}} & \multicolumn{2}{c}{\textsf{Named Entity Recognition}} \\
  \cmidrule(lr){2-3}  \cmidrule(lr){4-5} \cmidrule(lr){6-7}
  Embeddings & $\Delta$ F1 & $\Delta$ Recall & $\Delta$ F1  & $\Delta$ Recall  & $\Delta$ F1   & $\Delta$ Recall   \\
 \midrule
 Hard-Debias    & -0.657$\pm$\small{0.437}  & -1.220$\pm$\small{0.938}   & -0.007$\pm$\small{0.001}  & -0.025$\pm$\small{0.003}   & -0.004$\pm$\small{0.001}  & -0.015$\pm$\small{0.005} \\
 GN-Debias       & -0.594$\pm$\small{0.367} & -1.115$\pm$\small{0.821}   & -0.003$\pm$\small{0.001} & -0.010$\pm$\small{0.003}   & -0.002$\pm$ \small{0.001}  & -0.008$\pm$\small{0.002} \\
 ATT-Debias   & -0.689$\pm$\small{0.474} & -1.279$\pm$\small{1.000} & -0.024$\pm$\small{0.005} & -0.091$\pm$\small{0.019} & -0.013$\pm$\small{0.003} & -0.046$\pm$\small{0.011} \\
 CPT-Debias   & -0.501$\pm$\small{0.277} & -0.959$\pm$\small{0.674} & -0.004$\pm$\small{0.001} & -0.016$\pm$\small{0.005} & -0.002$\pm$\small{0.000} & -0.008$\pm$\small{0.001}  \\
 AE-Debias     & -2.862$\pm$\small{1.632} & -8.647$\pm$\small{5.072}   & -2.108$\pm$\small{0.558}  & -7.753$\pm$\small{1.996}   & -1.669$\pm$\small{0.547}  & -5.895$\pm$\small{1.893} \\
 AE-GN-Debias  & -3.505$\pm$\small{1.498}  & -10.766$\pm$\small{4.525}  & -4.765$\pm$\small{0.402}  & -16.760$\pm$\small{1.299} & -4.460$\pm$\small{0.485} & -5.097$\pm$\small{1.524} \\
 GP-Debias     & -2.911$\pm$\small{1.664}  & -8.810$\pm$\small{5.156}  & -2.058$\pm$\small{0.555} & -7.573$\pm$\small{1.988}  & -1.611$\pm$\small{0.542} & -5.696$\pm$\small{1.877} \\ 
 GP-GN-Debias  & -3.560$\pm$\small{1.506}  & -10.943$\pm$\small{4.557}  & -4.791$\pm$\small{0.391}  & -16.843$\pm$\small{1.262}  & -4.485$\pm$\small{0.468}  & -5.176$\pm$\small{1.471} \\
 \midrule
 CF-Debias  &-0.327$\pm$\small{0.248}	  & -0.621$\pm$\small{0.564}   & $\mathbf{0.000}\pm$\small{0.000}   &  $\mathbf{-0.001}\pm$\small{0.001} & $\mathbf{0.000}\pm$\small{0.000}  & $\mathbf{-0.001}\pm$\small{0.001}  \\
 CF-Debias-LA & -0.287$\pm$\small{0.118} & -0.506$\pm$\small{0.260}   & -0.002$\pm$\small{0.001}  & -0.006$\pm$\small{0.004} & -0.002$\pm$\small{0.001}  & -0.007$\pm$\small{0.005}  \\
 CF-Debias-KA  & $\mathbf{-0.123}\pm$\small{0.135} & $\mathbf{-0.186}\pm$\small{0.208} & $\mathbf{0.000}\pm$\small{0.000}   &  $\mathbf{-0.001}\pm$\small{0.001} & $\mathbf{0.000}\pm$\small{0.000}  & $\mathbf{-0.001}\pm$\small{0.001}  \\
 \bottomrule
 \end{tabular}
 \end{adjustbox}
 \caption{Performance degradation percentage with standard deviation for downstream tasks of POS Tagging, POS Chunking, and NER. The best model is indicated as boldface.}
 \label{table:down}
\end{table*}
\noindent To demonstrate the indirect gender bias in the word embedding, we perform two qualitative analyses from \citet{gonen19}. We take the top 500 male-biased words and the top 500 female-biased words, which becomes a word collection of the top 500 and the bottom 500 inner product between the word embeddings and $\overrightarrow{he}-\overrightarrow{she}$. From the debiasing perspective, these 1,000 word vectors should not be clustered distinctly. Therefore, we create two clusters with K-means and check the heterogeneity of the clusters through the cluster majority classification. The left side on Figure \ref{fig:nearest} shows that CF-Debias-KA generates a gender-invariant embedding for gender-biased wordsets by showing the lowest cluster classification accuracy. \\
\indent \citet{gonen19} demonstrates that the original bias\footnote{the dot-product between the original word embedding from GloVe and $\overrightarrow{he}-\overrightarrow{she}$} has a high correlation with the male/female ratio of the gender-biased words among the nearest neighbors of the word embedding. The right side of Figure \ref{fig:nearest}\footnote{Full plots of other baselines for two qualitative analyses are available in Appendix E and F,  respectively.} shows each profession word at (the dot-product, the male/female ratio). CF-Debias-KA shows the minimal Pearson correlation coefficient between the two axes.

\subsection{Downstream Task of Debiased Word Embeddings}
\noindent 
We compared multiple downstream task performances of the original and the debiased word embeddings, to check the ability to preserve semantic information in debiasing procedures. Following CoNLL 2003 shared task \cite{sang2002}, we selected Part-Of-Speech tagging, Part-Of-Speech chunking, and Named Entity Resolution as our tasks.
Table \ref{table:down} shows that there are constant performance degradation effects for all debiasing methods from the original embedding. However, our methods minimized the degradation of performances across the baseline models. Especially, CF-Debias-KA shows the minimal performance degradations by utilizing the nonlinear alignment regularization.

\begin{figure}[hbt!]
  \begin{subfigure}{0.58\columnwidth}
  \includegraphics[width=\linewidth,height=1in]{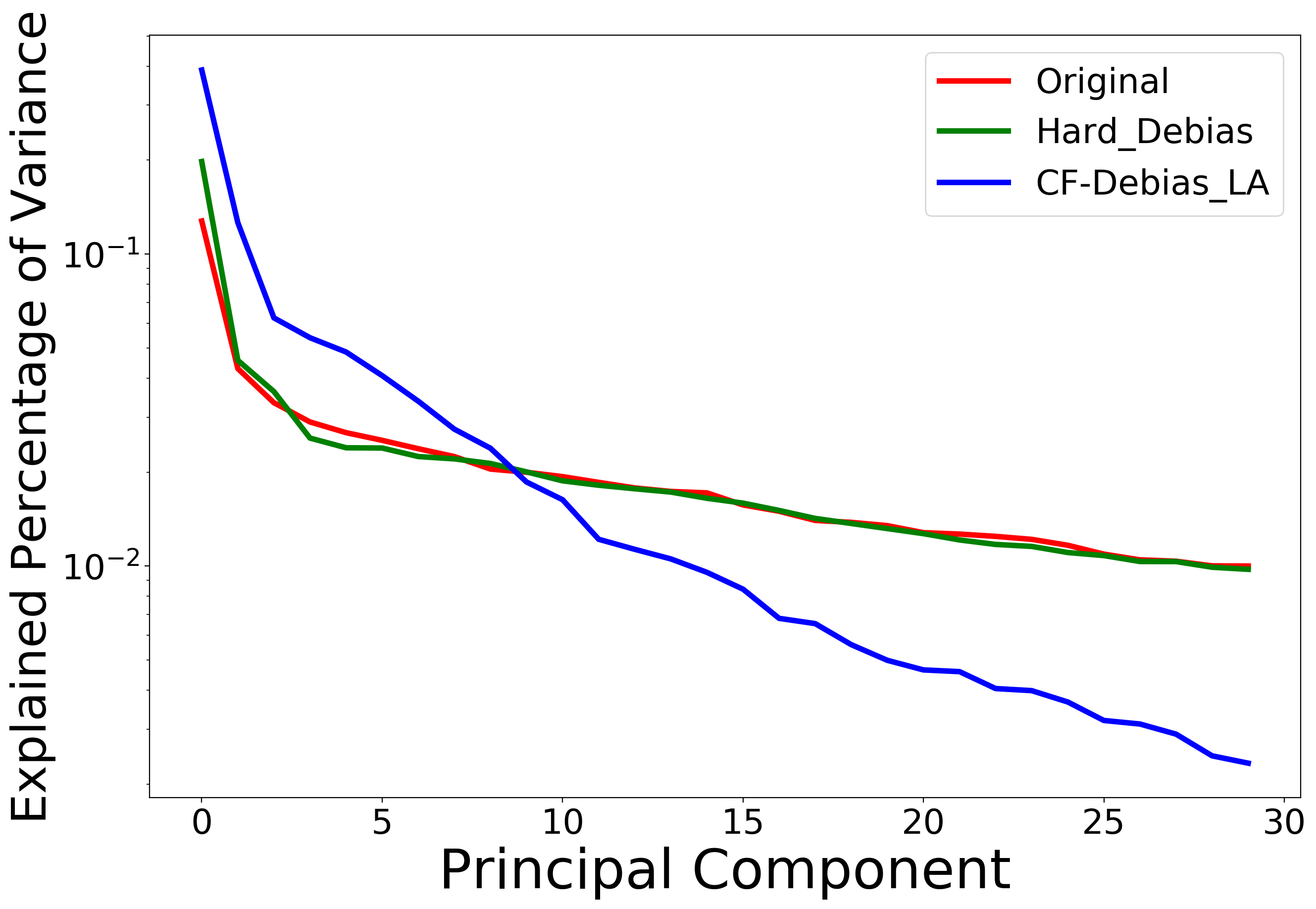}
  \end{subfigure}
  \hfill
  \begin{subfigure}{0.4\columnwidth}
  \includegraphics[width=\linewidth,height=1in]{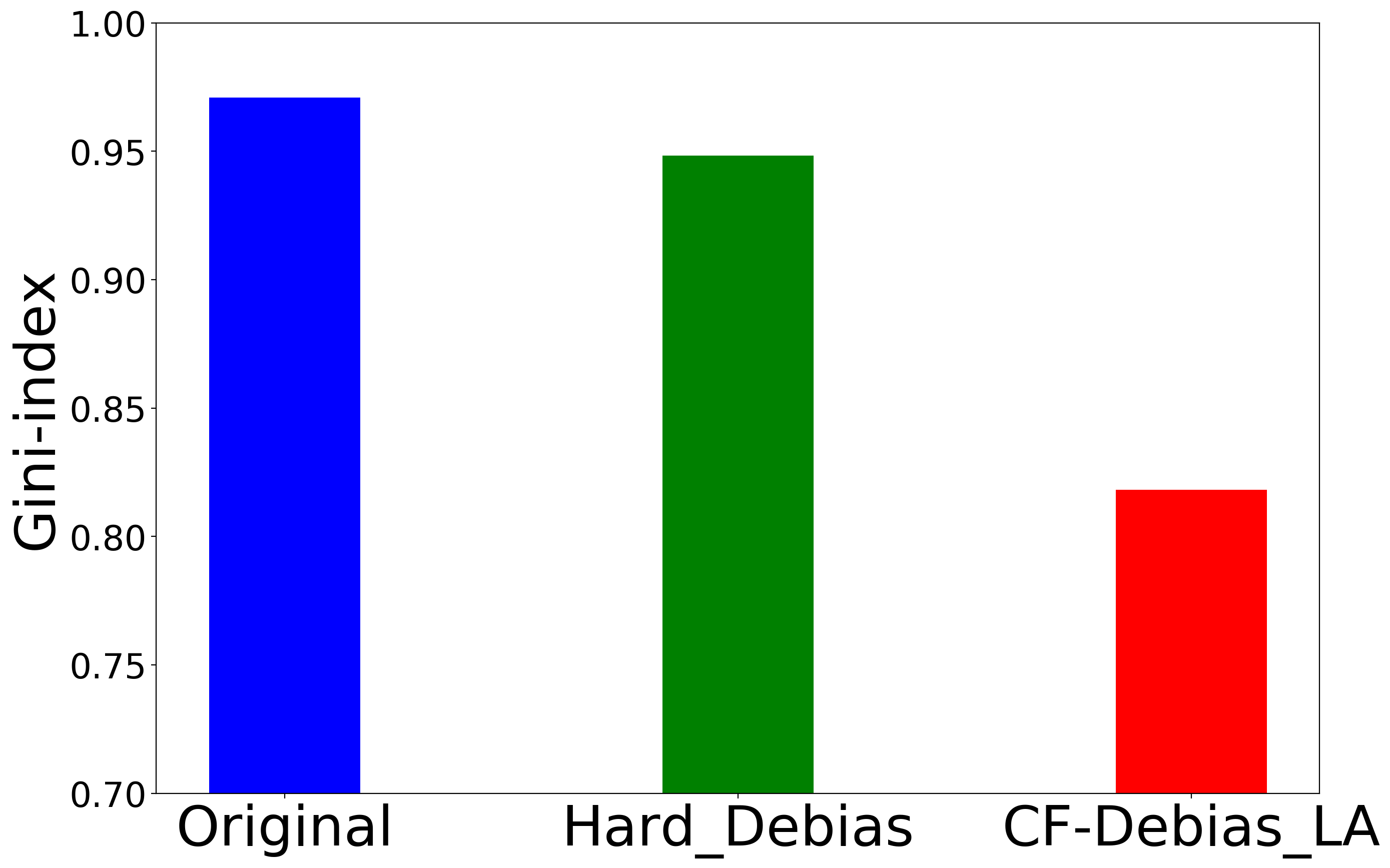}
  \end{subfigure} 
  \caption{The proportion (Left) and Gini-index (Right) from the variance vector for top 30 $PCs$ of difference vectors for gender word pairs}
  \label{fig:linear}
  \end{figure}

\subsection{Analyses on Alignment Regularization}
\noindent If the difference vectors of gender word pairs are not linearly aligned, the gender direction vector $\boldsymbol{v}_{g}$ in Eq. \eqref{eq:gender_direction} cannot be a pure direction of the gender information. Hence, we compared the variances explained by the top 30 principal components ($PC$) of difference vectors for gender word pairs, as a measurement for the linear alignment. The left plot in Figure \ref{fig:linear} shows the proportion of variances from each $PC$. Our method shows the largest concentration of the variances on a few components, other than Hard-Debias and Original embedding. The right plot in Figure \ref{fig:linear} shows Gini-index \cite{gini1912variabilita} for the variance proportion vector from $PCs$. Our method shows minimal Gini-index, which indicates the monopolized proportion of variances. 

\begin{figure}[hbt!]
  \begin{subfigure}{0.95\columnwidth}

  \includegraphics[height=1.5in]{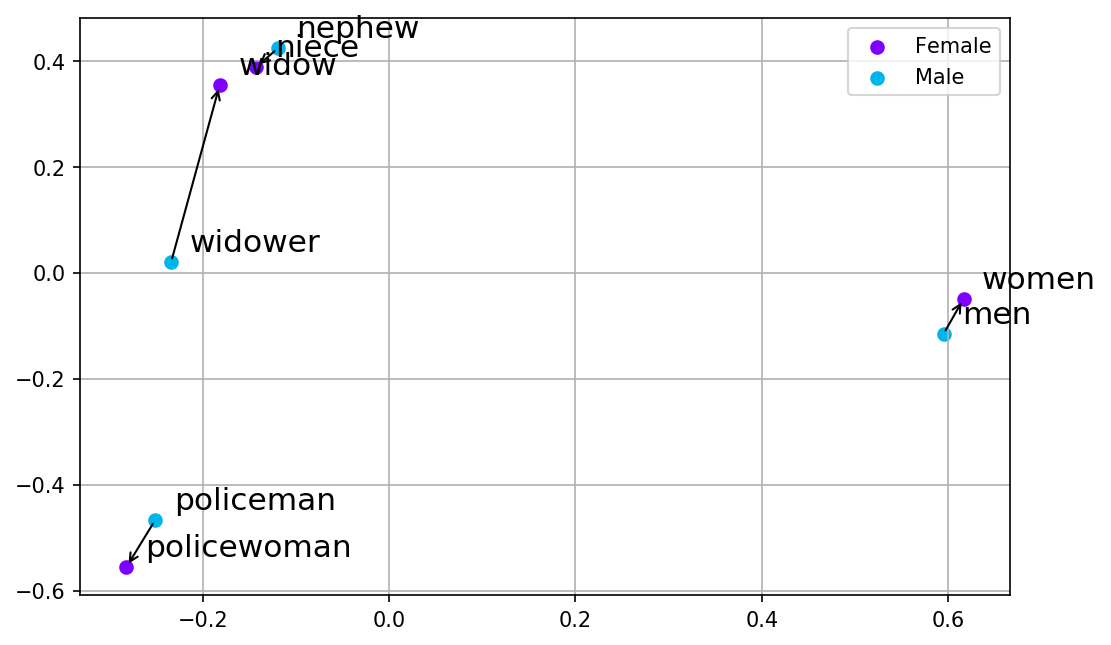}
  \end{subfigure}
  \begin{subfigure}{0.95\columnwidth}
  \includegraphics[height=1.5in]{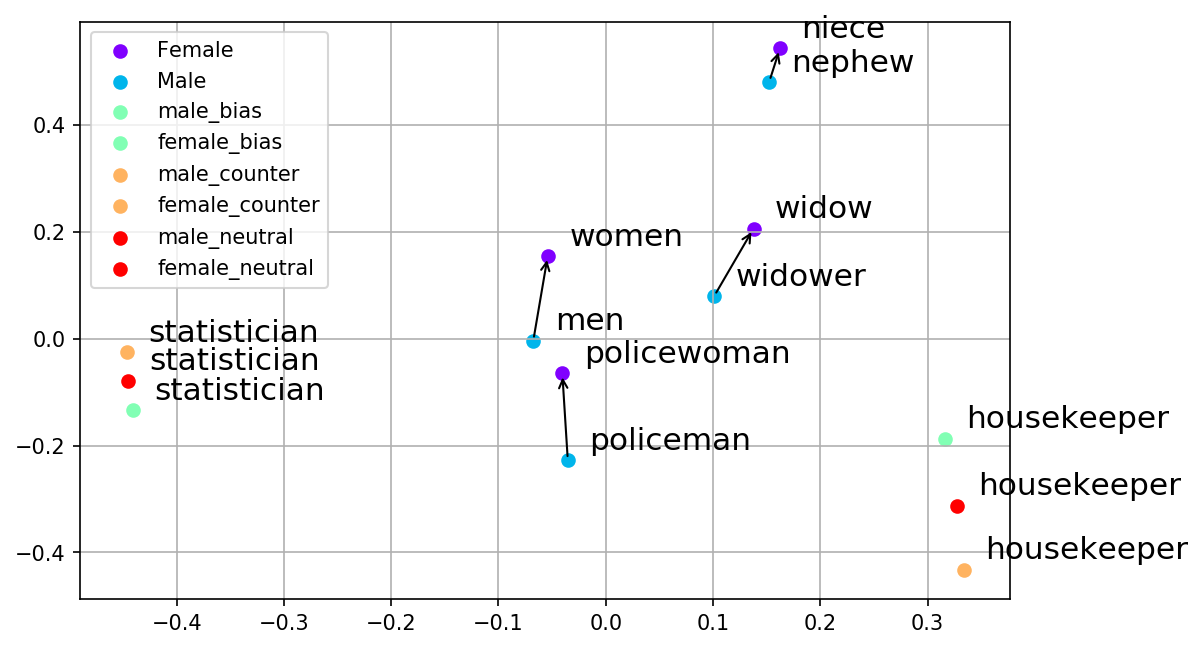}
  \end{subfigure} 
  \caption{LLE projection view of selected gender word pairs and biased word for original embedding space (left) and debiased embedding space (right)}
  \label{fig:align_view}
  \end{figure}

Also, Figure \ref{fig:align_view} shows two example plots of a selected gender word pairs in the original embedding space (Upper) and the CF-Debias-LA embedding space (Lower), by Locally Linear Embedding (LLE), \cite{roweis2000}. The lower plot in Figure \ref{fig:align_view} shows the consistency of the gender direction, and the plot visually describes the neutralization of \textit{housekeeper, statistician} by utilizing the counterfactually augmented word embeddings.
\section{Conclusions}
\noindent This work contributes to natural language processing society in two folds. For gender debiasing application, our model produces the debiased embeddings that has the most neutral gender latent information as well as the efficiently maintained semantics for the various NLP downstream tasks. For methodological modeling, CF-Debias suggests a new method of disentangling the latent information of word embeddings with the gradient reversal layer and creating the counterfactual embeddings by exploiting the geometry of the embedding space. It should be noted that these types of latent modeling can be applied to diverse natural language tasks to control expressions on emotions, prejudices, ideologies, etc.

\section*{Acknowledgments}
This research was supported by Basic Science Research Program through the National Research Foundation of Korea (NRF) funded by the Ministry of Education (NRF-2018R1C1B6008652)

\bibliography{anthology,emnlp2020}
\bibliographystyle{acl_natbib}

\newpage
\appendix

\section{The Derivation of Principal Component on Kernelized Alignment}
\noindent Let's assume that we want to align a word embedding ${\boldsymbol{w}'}$ to the set of the word embeddings $\{\boldsymbol{w}_{i}\}_{i=1}^{N}$. Then, we introduce nonlinear mapping function $f$, which projects a word embedding $\boldsymbol{w}_{i} \in \mathbb{R}^{d}$ into a newly introduced feature space, $f(\boldsymbol{w}_{i}) \in \mathbb{R}^{m}$. If we assume that the mapped outputs from the word embeddings $\{f(\boldsymbol{w}_{i})\}_{i=1}^{N}$ are zero-centered, the covariance matrix can be estimated as follows:
\begin{align}
& \Sigma_{f} = \frac{1}{N}\sum^{N}_{i=1} f(\boldsymbol{w}_{i}) f({\boldsymbol{w}_{i}})^{T} \nonumber
\end{align}

Same as the main paper, we set $\phi_{k}$ and $\lambda_{k}$ to be $k$-th eigenvector and eigenvalue for the projected outputs of the given embeddings $\{f(\boldsymbol{w}_{i})\}_{i=1}^{N}$, respectively. Then, we can get following equation, which describes the eigen-decomposition of the covariance matrix.
\begin{align}
\Sigma_{f}\phi_{k} &= \frac{1}{N}\sum^{N}_{i=1} f(\boldsymbol{w}_{i}) f({\boldsymbol{w}_{i}})^{T}\phi_{k} \nonumber \\ &= \frac{1}{N}\sum^{N}_{i=1} (f(\boldsymbol{w}_{i}) \boldsymbol{\cdot} \phi_{k})f(\boldsymbol{w}_{i})  =  \lambda_{k}\phi_{k} \nonumber
\end{align}

From above function, $\phi_{k}$ can be represented as a linearly-weighted combination of the $N$ mapped outputs of word embeddings as follows:
\begin{align}
\phi_{k} &= \frac{1}{N\lambda_{k}}\sum^{N}_{i=1}(f(\boldsymbol{w}_{i}) \boldsymbol{\cdot} \phi_{k})f(\boldsymbol{w}_{i})  \nonumber
\end{align}

Then, we multiply  $f(\boldsymbol{w}_{j})$ for $j=1,...,N$ to both sides of the equation.
\begin{align}
 f(\boldsymbol{w}_{j}) \boldsymbol{\cdot} \phi_{k} &= \frac{1}{N\lambda_{k}}f(\boldsymbol{w}_{j})\sum^{N}_{i=1}(f(\boldsymbol{w}_{i}) \boldsymbol{\cdot} \phi_{k})f(\boldsymbol{w}_{i})  \nonumber \\
&= \sum^{N}_{i=1}\frac{1}{N\lambda_{k}}(f(\boldsymbol{w}_{i}) \boldsymbol{\cdot} \phi_{k})(f(\boldsymbol{w}_{i}) \boldsymbol{\cdot} f(\boldsymbol{w}_{j})) \nonumber 
\end{align}

We can replace an inner-product of the two mapped outputs,  $(f(\boldsymbol{w}_{i}) \boldsymbol{\cdot} f(\boldsymbol{w}_{j}))$, into kernel $\K(\boldsymbol{w}_{i},\boldsymbol{w}_{j})$, which represents an inner product of two vectors in the projected space, for the case when computing mapped results of given data is complex or impossible. 
\begin{align}
 f(\boldsymbol{w}_{j}) \boldsymbol{\cdot} \phi_{k} & = \sum^{N}_{i=1}\frac{1}{N\lambda_{k}}(f(\boldsymbol{w}_{i}) \boldsymbol{\cdot} \phi_{k})\K(\boldsymbol{w}_{i},\boldsymbol{w}_{j}) \nonumber 
\end{align}
By letting $a^{i}_{k} = \frac{1}{N\lambda_{k}}(f(\boldsymbol{w}_{i}) \boldsymbol{\cdot} \phi_{k})$, we get
\begin{align}
f(\boldsymbol{w}_{j}) \boldsymbol{\cdot} \phi_{k} = \lambda_{k}N a^{j}_{k} = \sum^{N}_{i=1}a^{j}_{k}\K(\boldsymbol{w}_{i},\boldsymbol{w}_{j})
\nonumber 
\end{align}
The above equation can be represented as the $j$-th component of the $k$-th eigenvector-decomposition problem of $\Ker$, which is a matrix of $N \times N$ kernel elements $\K(\boldsymbol{w}_{i},\boldsymbol{w}_{j})$ for $i,j = 1,...,N$. See the below equation, which is $k$-th eigenvector-decomposition problem of $\Ker$, when $\boldsymbol{a}_{k} = {[a^{1}_{k},...,a^{N}_{k}]}^{T}$.
\begin{align}
\lambda_{k}N\boldsymbol{a}_{k} = \Ker\boldsymbol{a}_{k}
\nonumber 
\end{align}
This implication means that  $a^{j}_{k}$ is $j$-th component of $k$-th eigenvector of $\Ker$ and we can compute $a^{j}_{k}$ by solving eigen-decomposition problem of $\Ker$.

Substituting $f(\boldsymbol{w}_{j})$ on above equation into $f(\boldsymbol{w}')$, which is mapped result of  the target word embedding $\boldsymbol{w}'$, we get ${PC}_{k}$, $k$-th principal component of new word embedding $\boldsymbol{w}'$ on the projected space as follows:
\begin{align}
\label{eq:principal_component}
{PC}_{k}=f(\boldsymbol{w}') \cdot \phi_{k} =  &\Sigma^{N}_{i=1}{a^{i}_{k}f(\boldsymbol{w}_{i}) \cdot f(\boldsymbol{w}')} \nonumber \\= &\Sigma^{N}_{i=1}{a^{i}_{k}\Ker(\boldsymbol{w}_{i},\boldsymbol{w}')}
\end{align}

It should be noted that above derivation is based on \citet{1998nonlinear}. The proposed Kernelized alignment can be seen as an example which applies an nonlinear alignment to the word embeddings, by utilizing the kernel trick provided from \citet{1998nonlinear}.

\newpage

\section{Notation table}
\begin{table}[hbt!]
\centering
\resizebox{1\linewidth}{!}{
\begin{tabular}{|c|l|}
\hline
\textbf{Notation} & \textbf{Description} \\
\hline
$\boldsymbol{{w}}_{f}$ & The embedding of \textit{feminine} word \\
\hline
$\boldsymbol{{w}}_{m}$ & The embedding of \textit{masculine} word \\
\hline
$\boldsymbol{{w}}_{n}$ & The embedding of \textit{gender neutral} word \\
\hline
$V_f$ & The \textit{feminine} word set \\
\hline
$V_m$ & The \textit{masculine} word set \\
\hline
$V_n$ & The \textit{gender neutral} word set \\
\hline
$\boldsymbol{{z}}_{f}^{s}$ & The semantic latent variable of $w_f$ \\
\hline
$\boldsymbol{{z}}_{m}^{s}$ & The semantic latent variable of $w_m$ \\
\hline
$\boldsymbol{{z}}_{n}^{s}$ & The semantic latent variable of $w_n$ \\
\hline
$\boldsymbol{{z}}_{f}^{g}$ & The gender latent variable of $w_f$ \\
\hline
$\boldsymbol{{z}}_{m}^{g}$ & The gender latent variable of $w_m$ \\
\hline
$\boldsymbol{{z}}_{n}^{g}$ & The gender latent variable of $w_n$ \\
\hline
$\neg\boldsymbol{{z}}_{n}^{g}$ & The counterfactual-gender latent variable \\
\hline
$\boldsymbol{{\hat{w}}}_{f}$ & The reconstructed word embedding of $\boldsymbol{{w}}_{f}$ \\
\hline
$\boldsymbol{{\hat{w}}}_{m}$ & The reconstructed word embedding of $\boldsymbol{{w}}_{m}$ \\
\hline
$\boldsymbol{{\hat{w}}}_{n}$ & The reconstructed word embedding of $\boldsymbol{{w}}_{n}$ \\
\hline
$\boldsymbol{{\hat{w}}}_{cf}$ & The counterfactually reconstructed word embedding \\
\hline
$\boldsymbol{{\hat{w}}}_{neu}$ & The gender neutralized word embedding \\
\hline
$g_f$ & The output of gender classifier for $z_f^g$\\
\hline
$g_m$ & The output of gender classifier for $z_m^g$ \\
\hline
$\boldsymbol{{v}}_{g}$ & The gender direction vector \\
\hline
$\Omega$ & The gender word pairs set \\
\hline
$E$ & The encoder of our method\\
\hline
$D$ & The decoder of our method\\
\hline
$C_r$ & The auxilary gender classifier \\
\hline
$C_a$ & The gender latent generator  \\
\hline
\end{tabular}
}
\caption{The description of the notations in this paper.
}
\label{table:notation}
\end{table}

\section{WEAT Hypothesis test}
\noindent WEAT hypothesis \cite{weat2017} test quantifies the bias with effect size and p-value.
We can compute the effect size of the two target words set against two attribute words set. To quantify the gender bias, we use \cite{weat2019} subset of masculine $(A_{1})$ and feminine words$(A_{2})$ as an attribute words, and use career $(T_{1})$ and family $(T_{2})$ related words for target words set. We compare the effect size and p-value for different experiment environment by changing the attribute words, as shown in Table 2 in the paper.

We can compute the association measure $s$, between target word $t$ and the attribute word set as follows:
\begin{align}
&s(t) = \frac{1}{|A_{1}|}\sum_{a_{1} \in A_{1}}\cos(t,a_{1}) - \frac{1}{|A_{2}|}\sum_{a_{2} \in A_{2}}\cos(t,a_{2}) \nonumber
\end{align}
We compute the effect size, the degree of bias, based on the difference between mean of association value as follows:
\begin{align}
&\frac{\text{Mean}_{t_{1}\in T_{1}}s(t_{1})-\text{Mean}_{t_{2}\in T_{2}}s(t_{2})}
{\text{std}_{t\in{T_{1} \cup T_{2}}}s(t)} \nonumber
\end{align}
To check the significant level of bias, we need to compute the test statistics, $s(T_{1},T_{2})$, and one-sided p-value. We compute the p-value based on $\{T_{1}^{(i)},T_{2}^{(i)}\}$, the all partition of $T_{1} \cup T_{2}$ as follows:
\begin{align}
s(T_{1},T_{2}) = \sum_{t_{1} \in T_{1}}&s(t_{1}) - \sum_{t_{2} \in T_{2}}s(t_{2}) \nonumber \\
\text{p-value}= P\{|s(T_{1}^{(i)},&T_{2}^{(i)})| > |s(T_{1},T_{2})|\} \nonumber
\end{align}

If the word embedding has a conventional gender bias, effective size can have a positive value, and negative value, otherwise.
To measure the gender bias properly, we need to consider both of conventional gender bias, and anti-conventional gender bias.
We compute the p-value based on the absolution value of test statistics to measure gender bias properly.

\section{Performance Test Result for Gender Classifier $C_{r}$}
\label{sec:genclass}
\begin{figure}[hbt!]
  \centering
  \includegraphics[width=2.6in,height=1.1in]{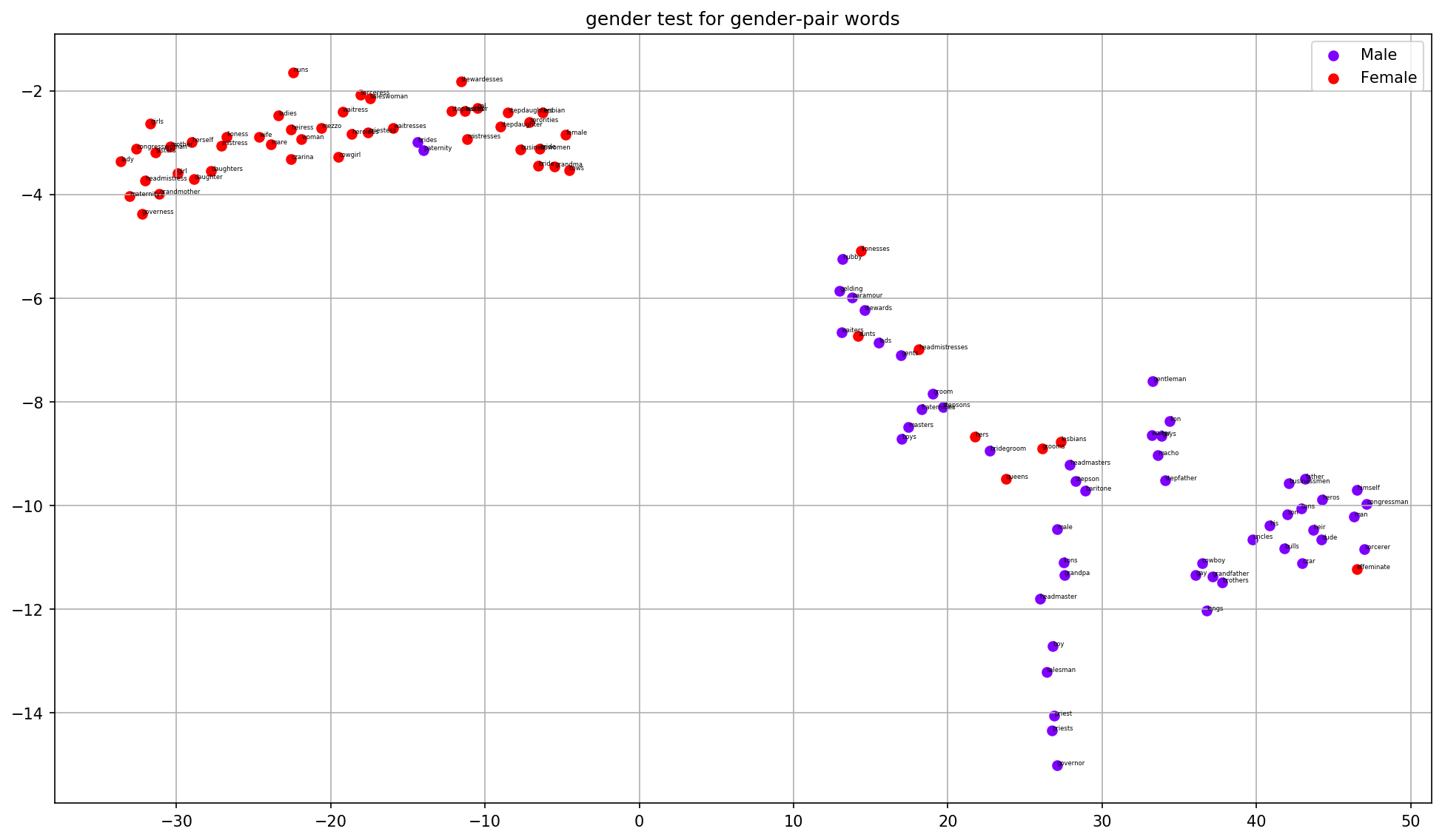}
  \caption{The t-SNE projection view of gender latent variables of the test gender word pairs}
  \label{fig:gender_latent}
  \end{figure}

\noindent To test gender indicating the ability of the gender classifier $C_r:\boldsymbol{z}^{g} \rightarrow [0,1]$, we tested indicating accuracy of the gender-definition words, i.e., he, she, etc.; and gender-stereotypical words, i.e., doctor, nurse, etc. We utilized 53 gender word pairs as test word pairs from entire gender word pairs, utilizing the remaining words for training. We selected well known gender-biased occupation words for examples of gender-stereotypical words, 10 for each gender case as follows: \\
\indent $[doctor, programmer, boss, maestro, warrior\\, john, politician, statistician, athlete, nurse, \\homemaker, cook, cosmetics, dancer, mary, \\violinist, housekeeper, secretary]$. \\
\indent The test accuracy for gender-definition words are 0.8490, 0.8867 for masculine and feminine words, respectively. For gender-stereotypical words, $C_{r}$ indicates correct gender biases for all male-biased words except the word \textit{athlete} and all female-biased words. Figure \ref{fig:gender_latent} shows the visual separation of gender latent variables for masculine words and feminine words.

\newpage
\onecolumn
\section{Full Plots for the Clustering Analysis}
\begin{figure*}[!htbp]
  \begin{subfigure}{0.49\linewidth}
  \centering
  \includegraphics[width=2.2in,height=1.2in]{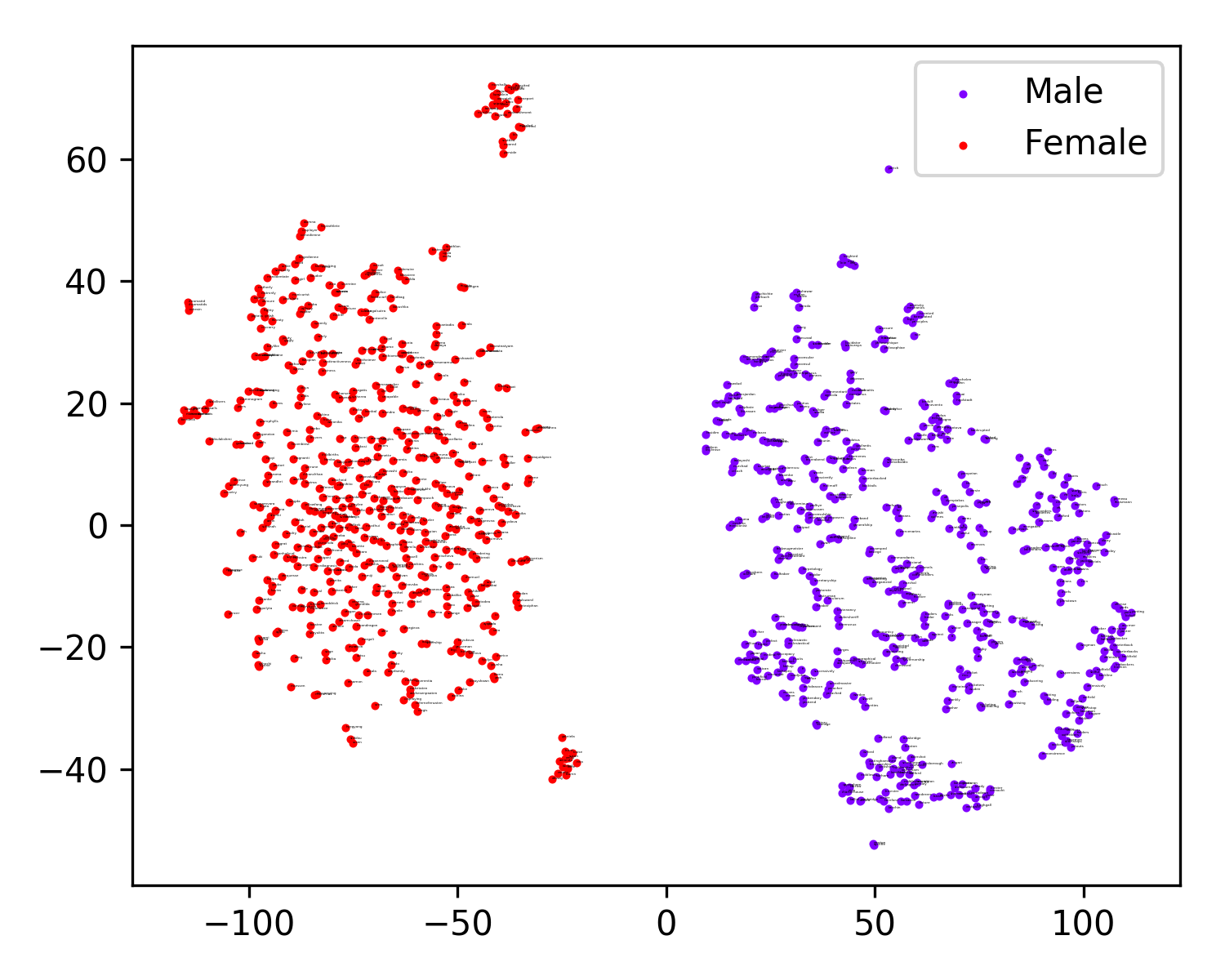}
  \caption{Original (99.9\%)}
  \end{subfigure}
  \hfill
  \begin{subfigure}{0.49\linewidth}
  \centering
  \includegraphics[width=2.2in,height=1.2in]{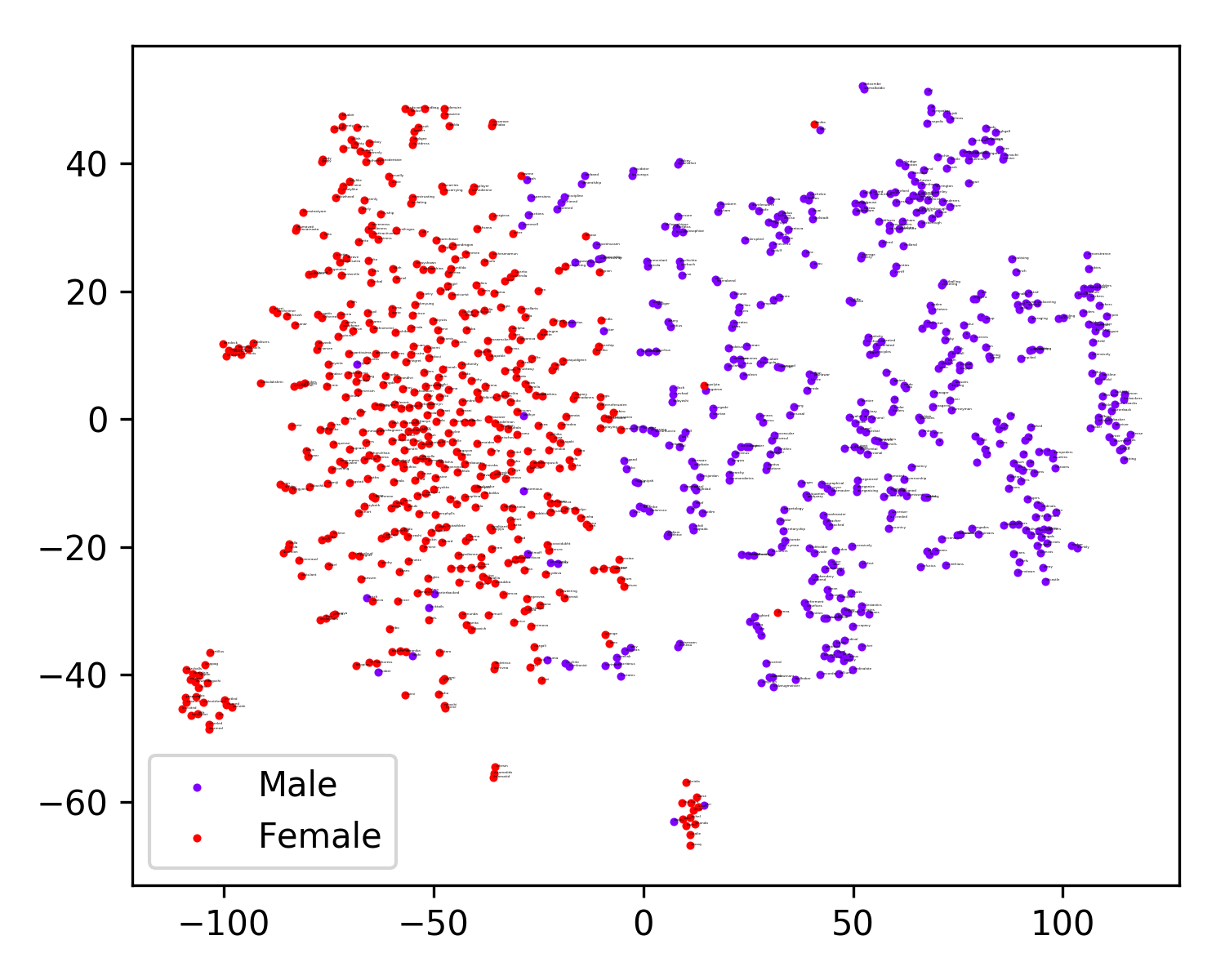}
  \caption{Hard-Debias (89.4\%)} 
  \end{subfigure} 
  \begin{subfigure}{0.49\linewidth} 
  \centering
  \includegraphics[width=2.2in,height=1.2in]{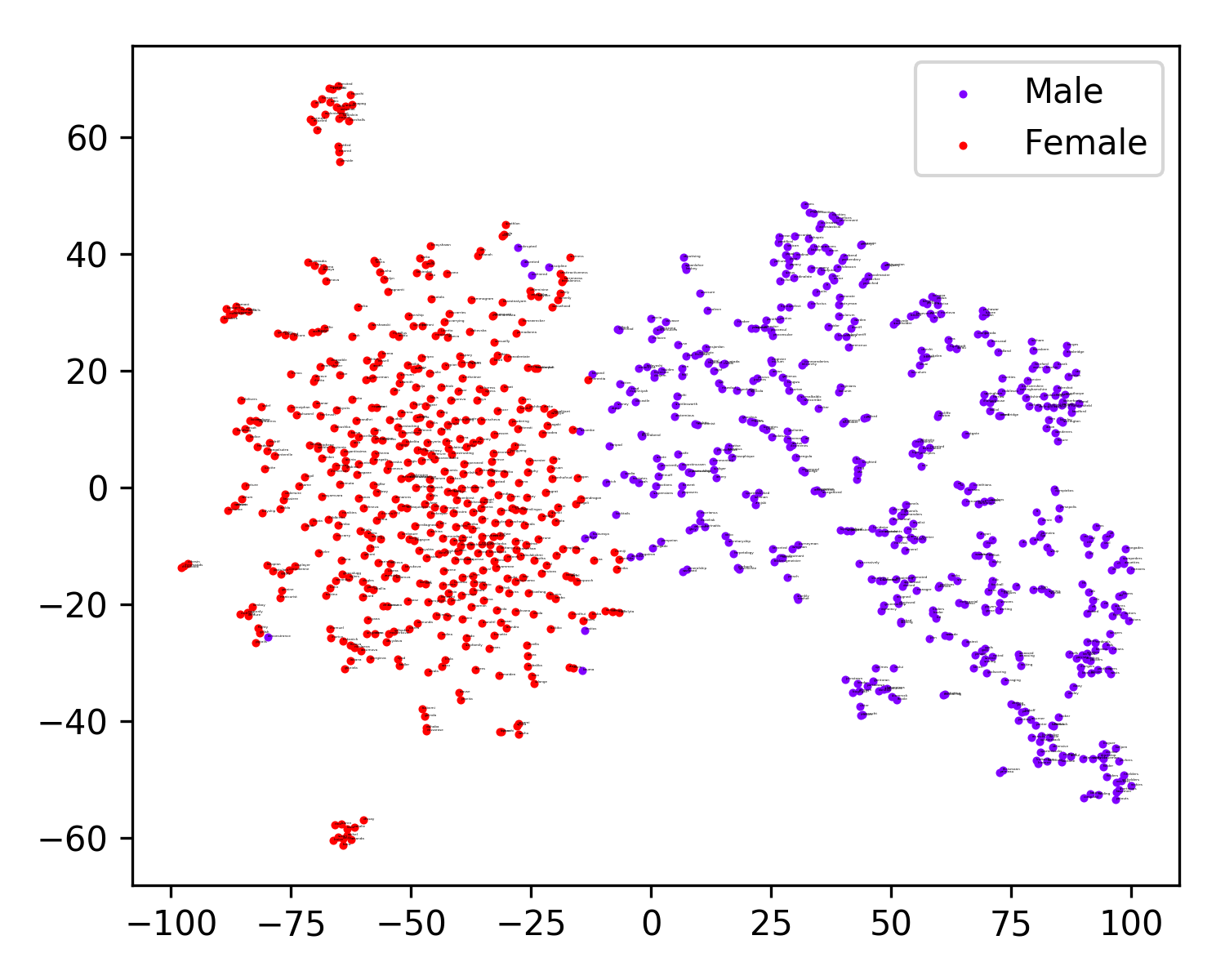} 
  \caption{GN-Debias (97.3\%)} 
  \end{subfigure}  
  \hfill 
  \begin{subfigure}{0.49\linewidth} 
  \centering
  \includegraphics[width=2.2in,height=1.2in]{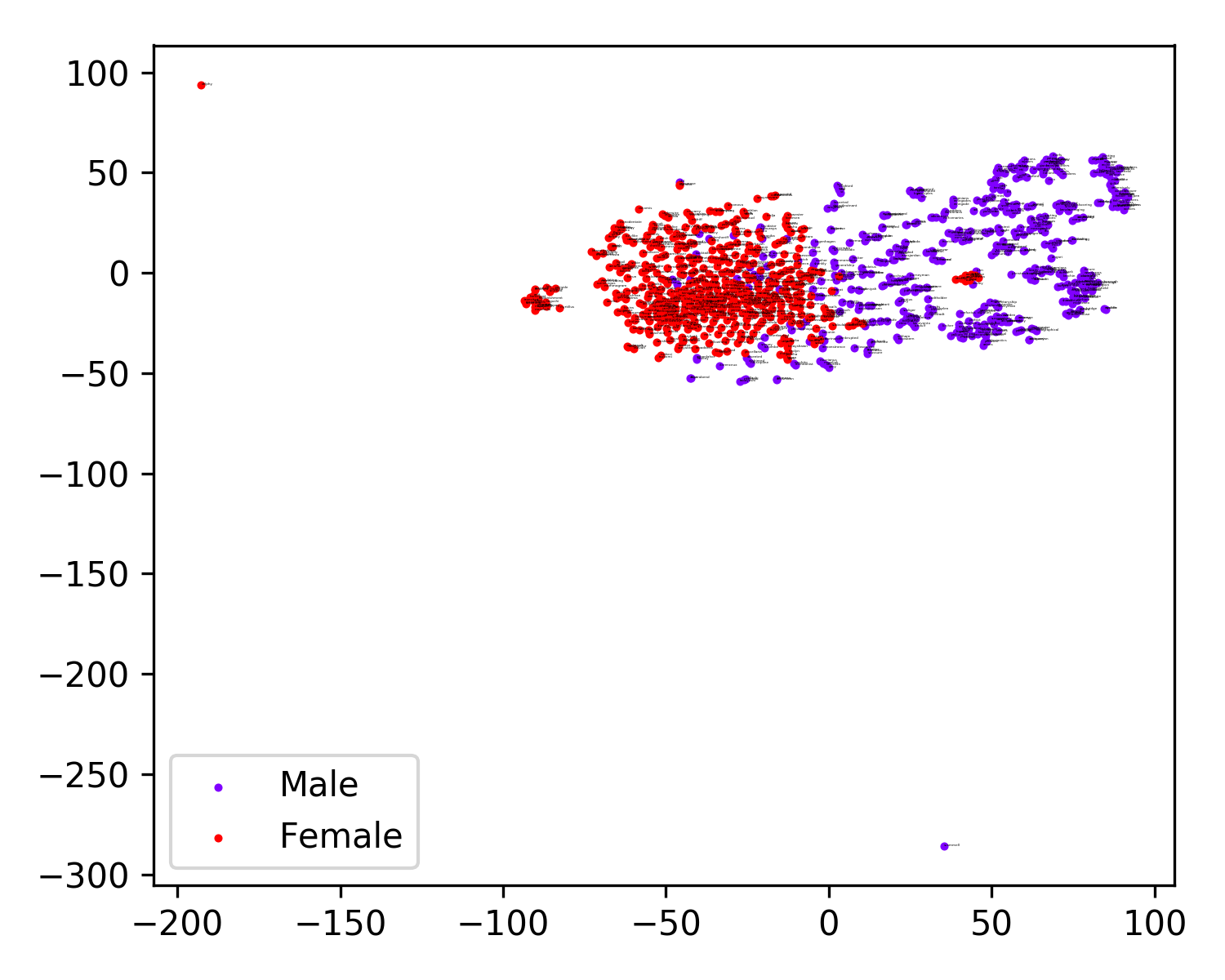} 
  \caption{ATT-Debias (99.1\%)} 
  \end{subfigure}
  \begin{subfigure}{0.49\linewidth} 
  \centering
  \includegraphics[width=2.2in,height=1.2in]{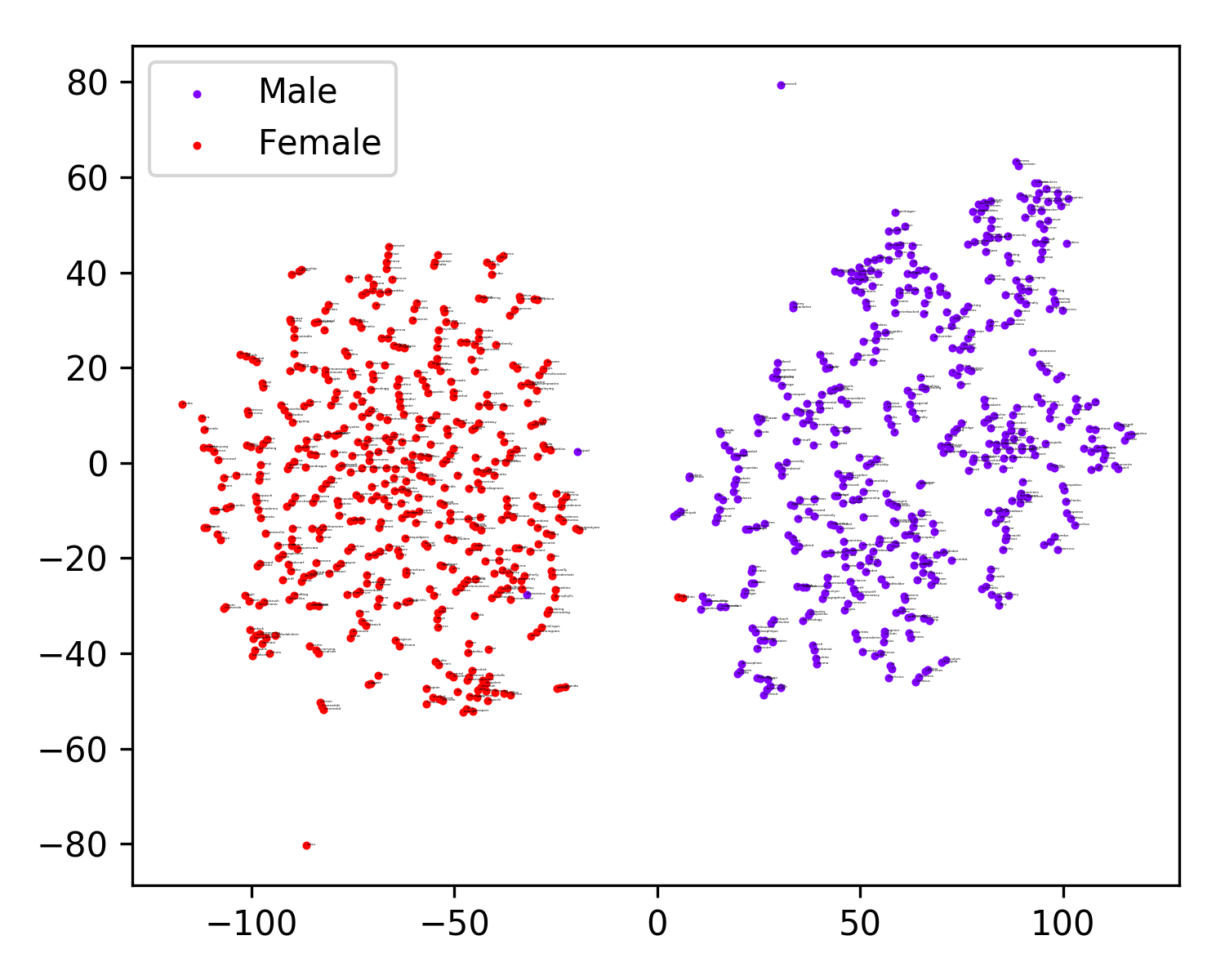} 
  \caption{CPT-Debias (100.0\%)} 
  \end{subfigure}  
  \hfill 
  \begin{subfigure}{0.49\linewidth} 
  \centering
  \includegraphics[width=2.2in,height=1.2in]{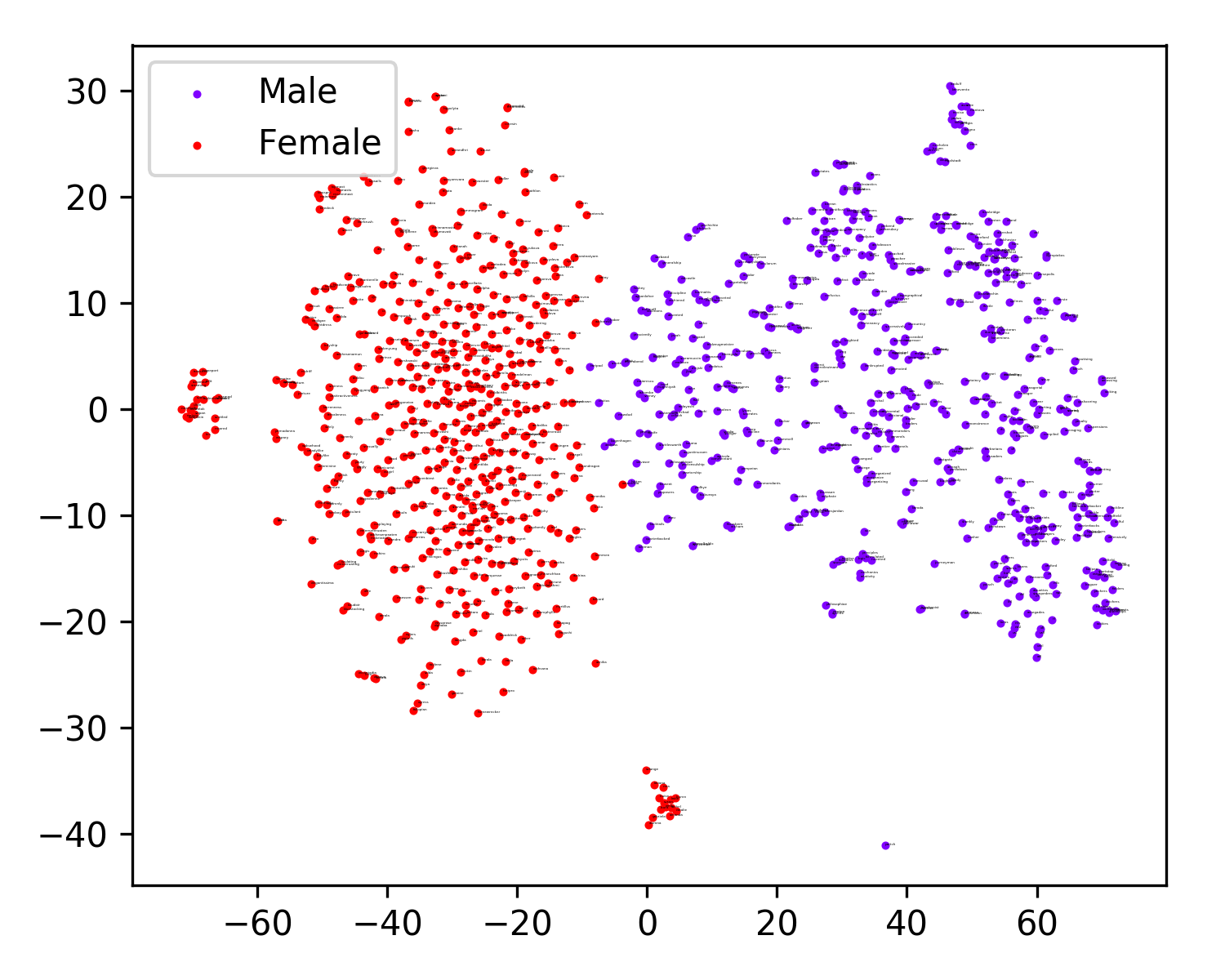} 
  \caption{AE-Debias (92.6\%)} 
  \end{subfigure}
  \begin{subfigure}{0.49\linewidth} 
  \centering
  \includegraphics[width=2.2in,height=1.2in]{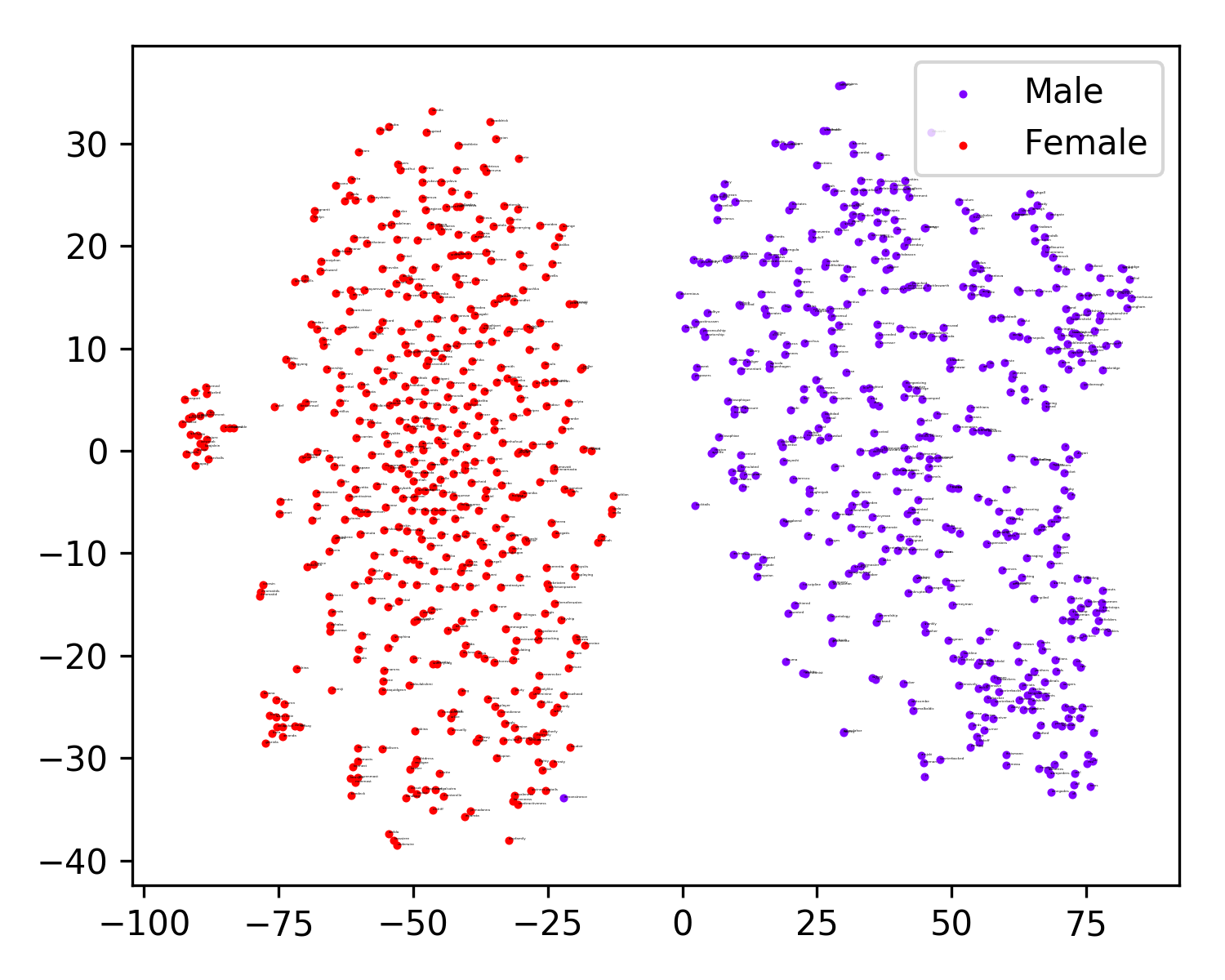} 
  \caption{AE-GN-Debias (100.0\%)} 
  \end{subfigure}  
  \hfill 
  \begin{subfigure}{0.49\linewidth} 
  \centering
  \includegraphics[width=2.2in,height=1.2in]{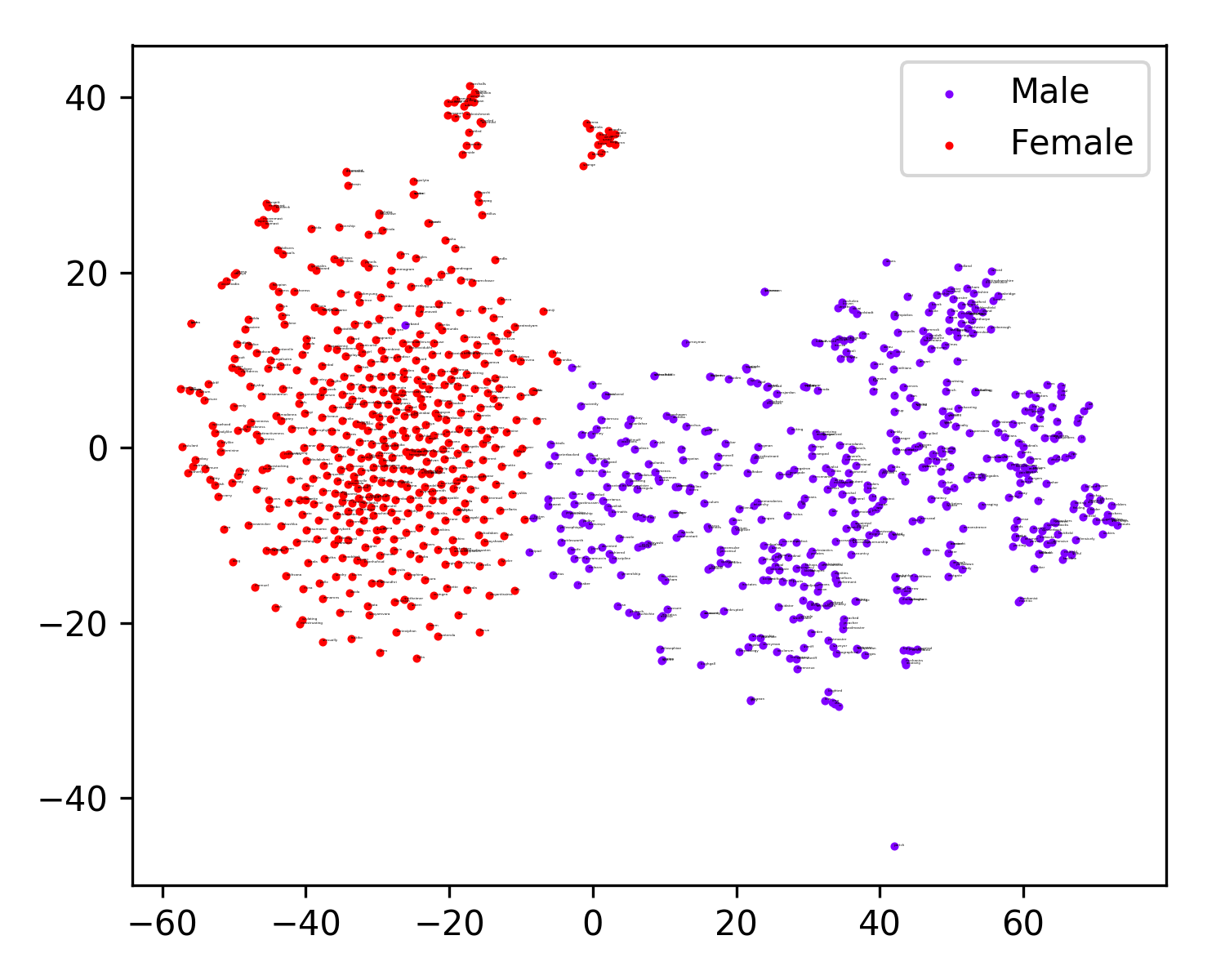} 
  \caption{GP-Debias (92.5\%)} 
  \end{subfigure}
  \begin{subfigure}{0.49\linewidth} 
  \centering
  \includegraphics[width=2.2in,height=1.2in]{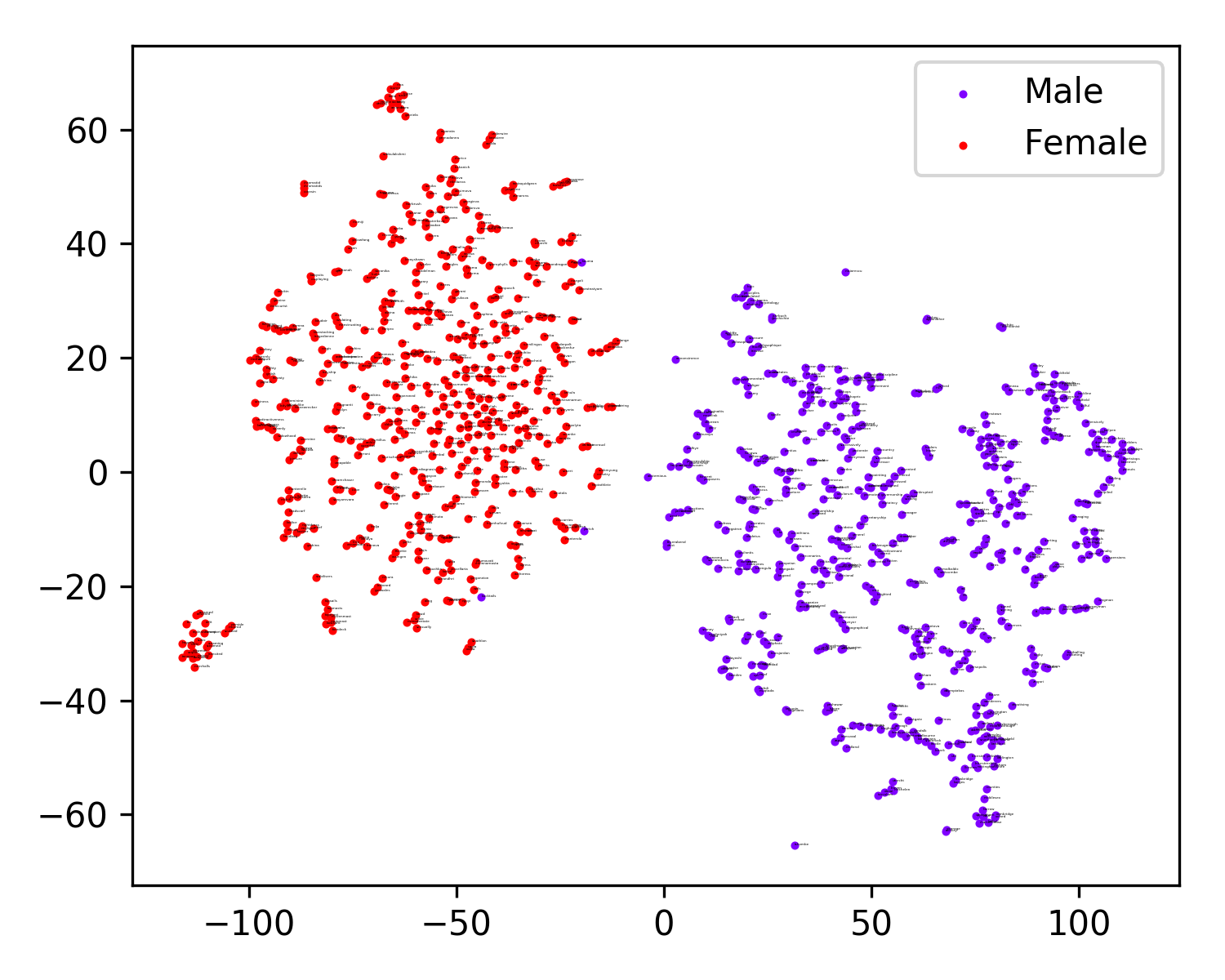} 
  \caption{GP-GN-Debias (100.0\%)} 
  \end{subfigure}  
  \hfill 
  \begin{subfigure}{0.49\linewidth} 
  \centering
  \includegraphics[width=2.2in,height=1.2in]{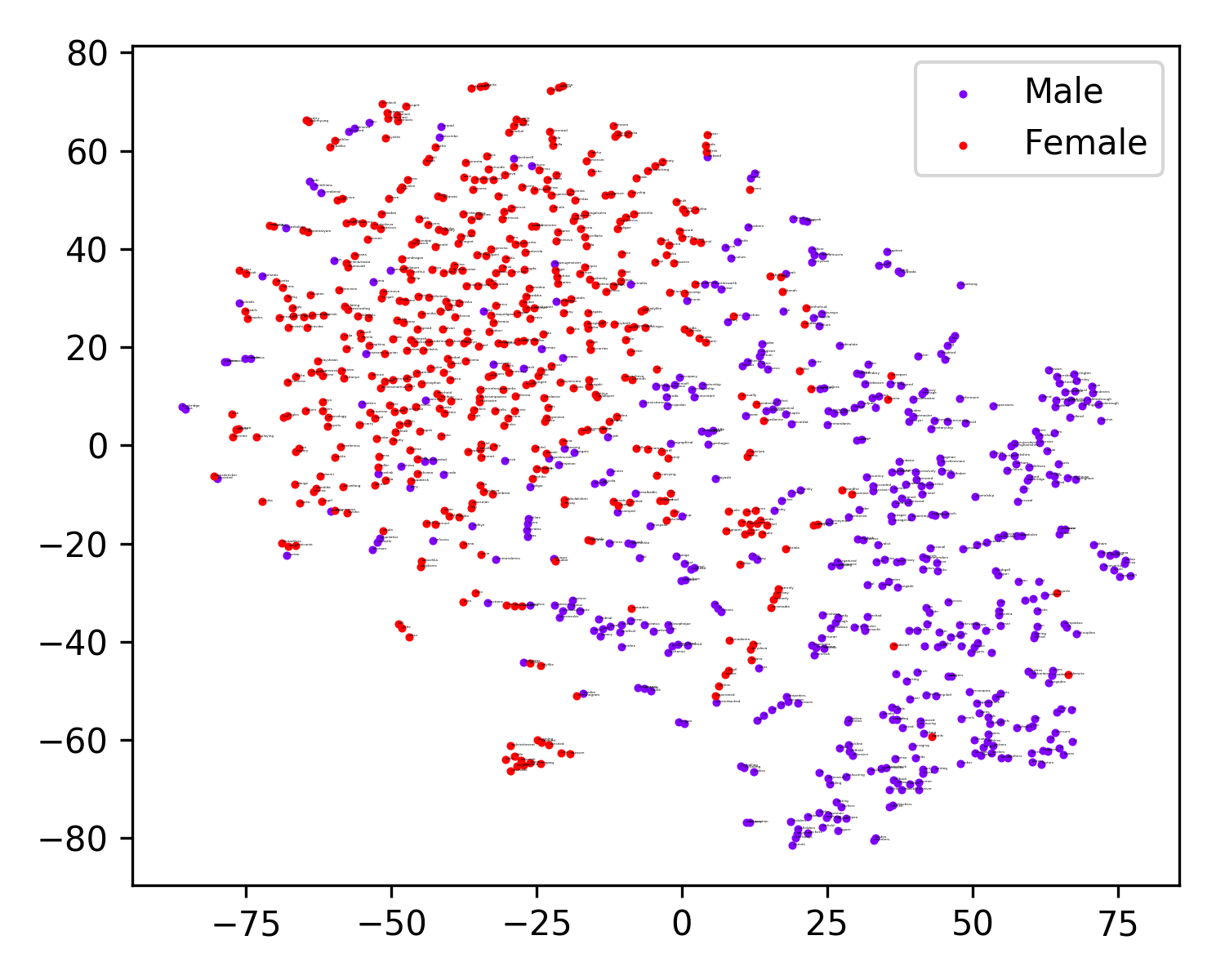} 
  \caption{CF-Debias (78.1\%)} 
  \end{subfigure}
  \begin{subfigure}{0.49\linewidth} 
  \centering
  \includegraphics[width=2.2in,height=1.2in]{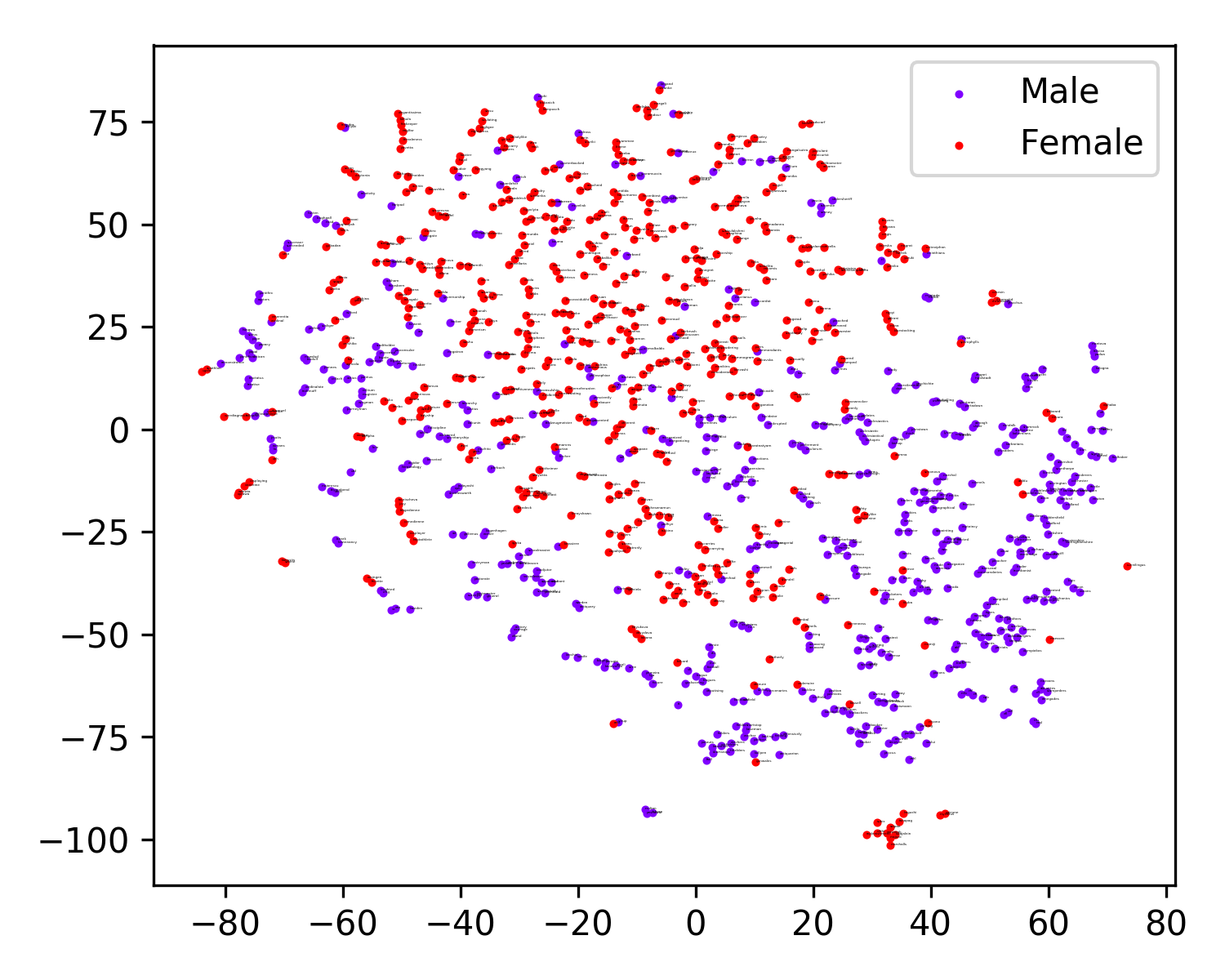} 
  \caption{CF-Debias-LA (63.1\%)} 
  \end{subfigure}  
  \hfill 
  \begin{subfigure}{0.49\linewidth} 
  \centering
  \includegraphics[width=2.2in,height=1.2in]{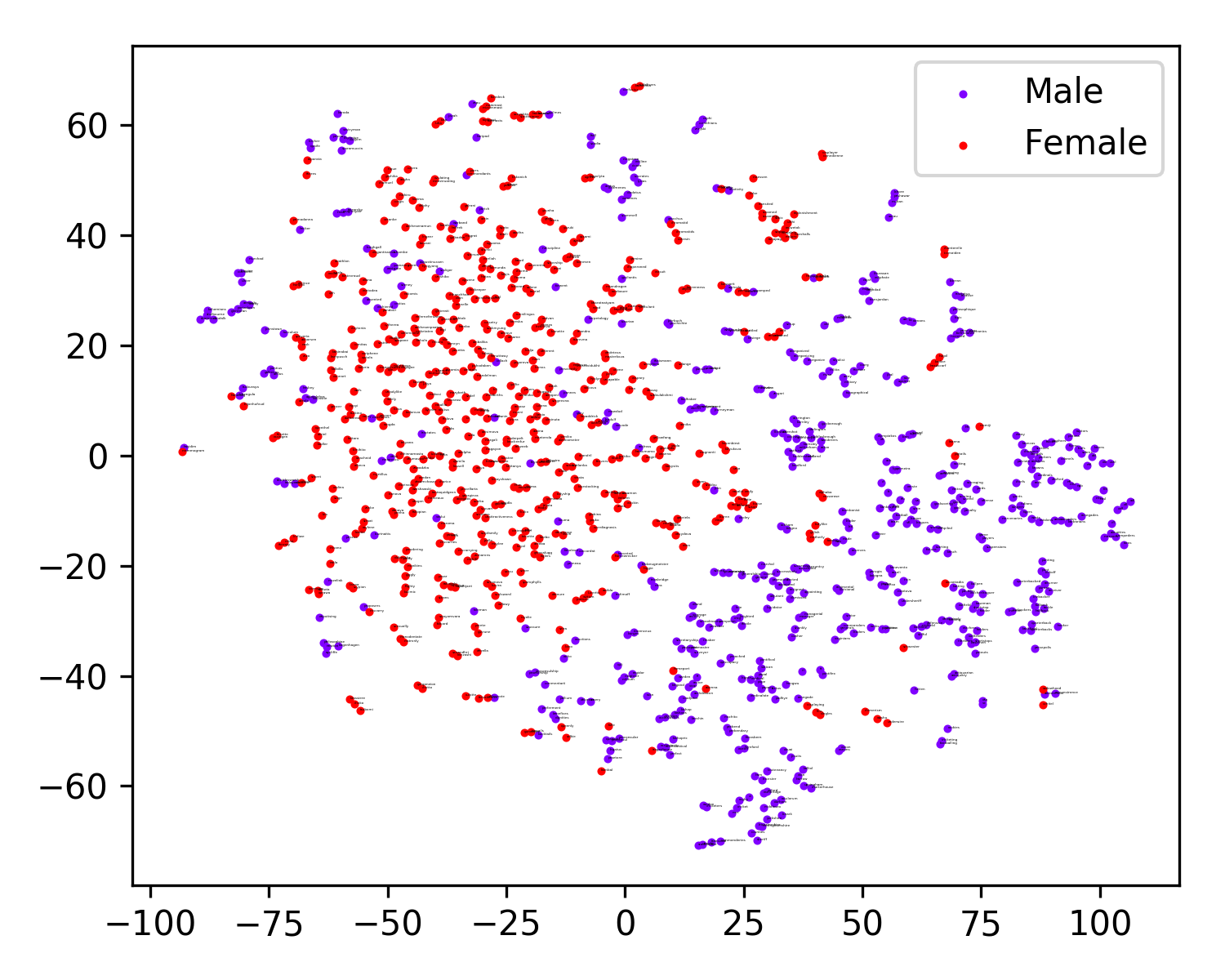} 
  \caption{CF-Debias-KA (76.8\%)} 
  \end{subfigure}
  \caption{The t-SNE projection views for embeddings of 500 male-biased words and 500 female-biased words according to the original Glove, the cluster majority based classification accuracy is added in parenthesis.}
  \label{fig:cluster}
  \end{figure*}
\FloatBarrier

\newpage
\onecolumn
\section{Full Plots for Correlation Analysis between Original Bias and Nearest Neighbors}
\begin{figure}[!htbp]
  \begin{subfigure}{0.49\linewidth}
  \centering
  \includegraphics[width=2.2in,height=1.2in]{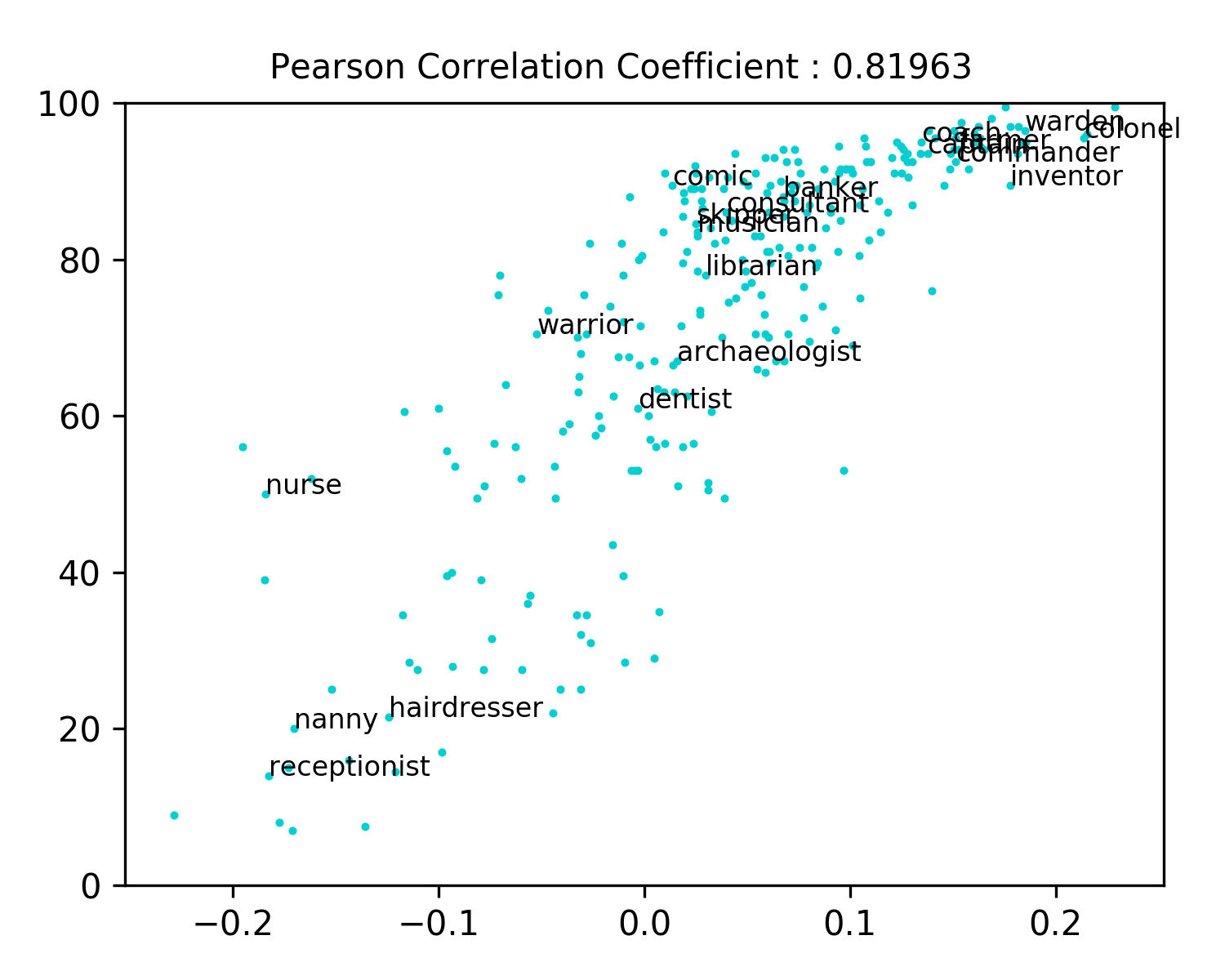}
  \caption{Original (0.8196)}
  \end{subfigure}
  \hfill
  \begin{subfigure}{0.49\linewidth}
  \centering
  \includegraphics[width=2.2in,height=1.2in]{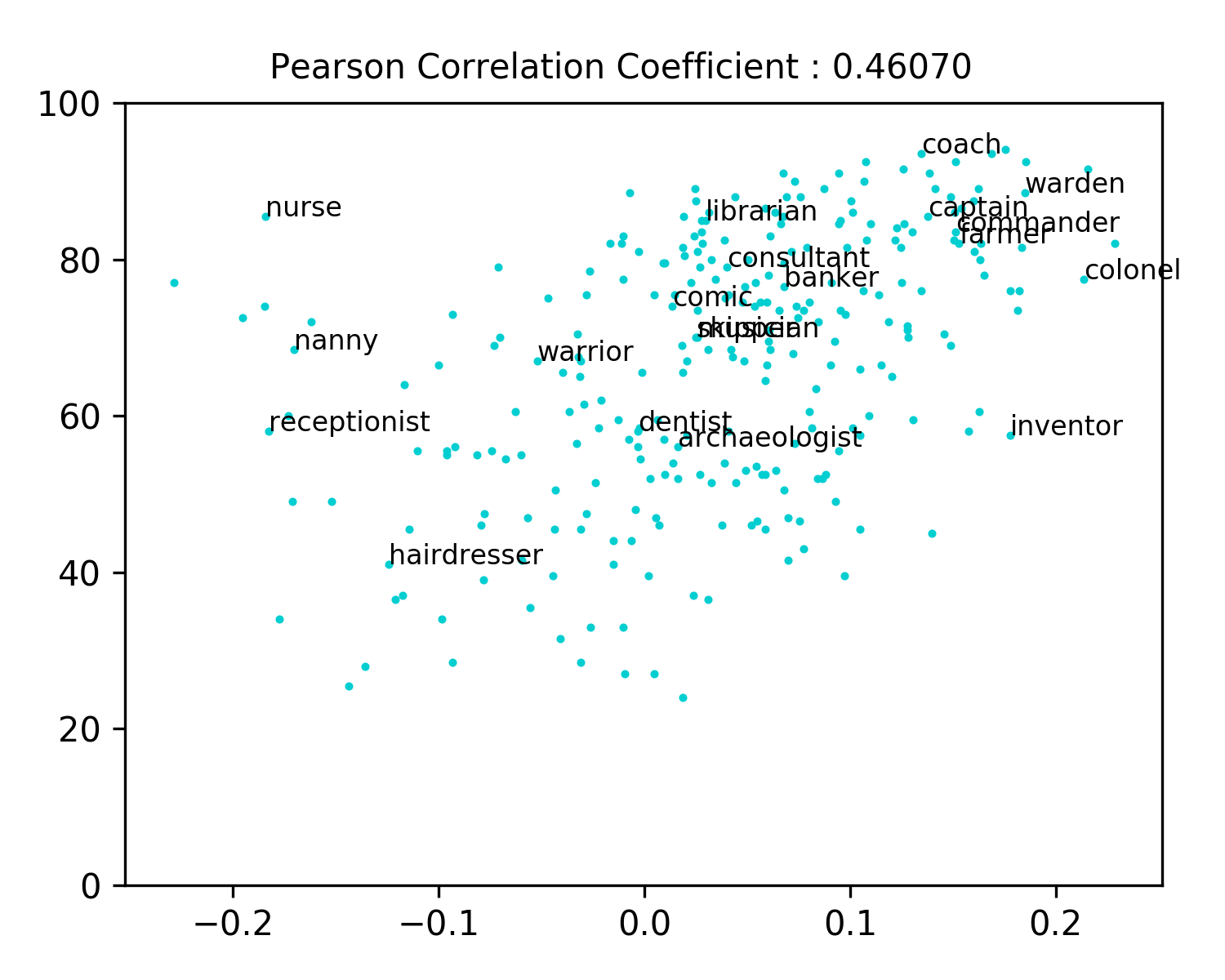}
  \caption{Hard-Debias (0.4607)} 
  \end{subfigure} 
  \begin{subfigure}{0.49\linewidth} 
  \centering
  \includegraphics[width=2.2in,height=1.2in]{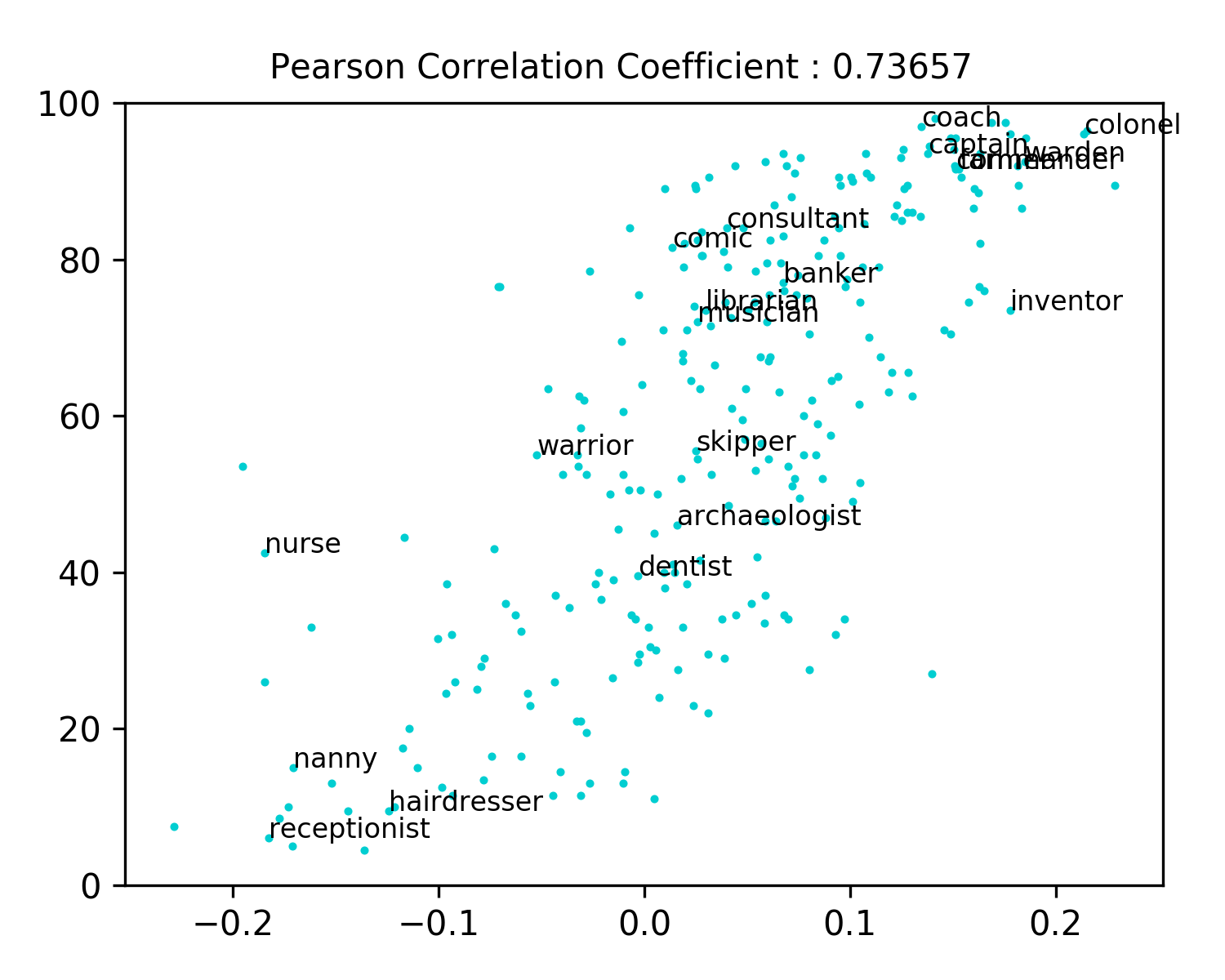} 
  \caption{GN-Debias (0.7366)} 
  \end{subfigure}  
  \hfill 
  \begin{subfigure}{0.49\linewidth} 
  \centering
  \includegraphics[width=2.2in,height=1.2in]{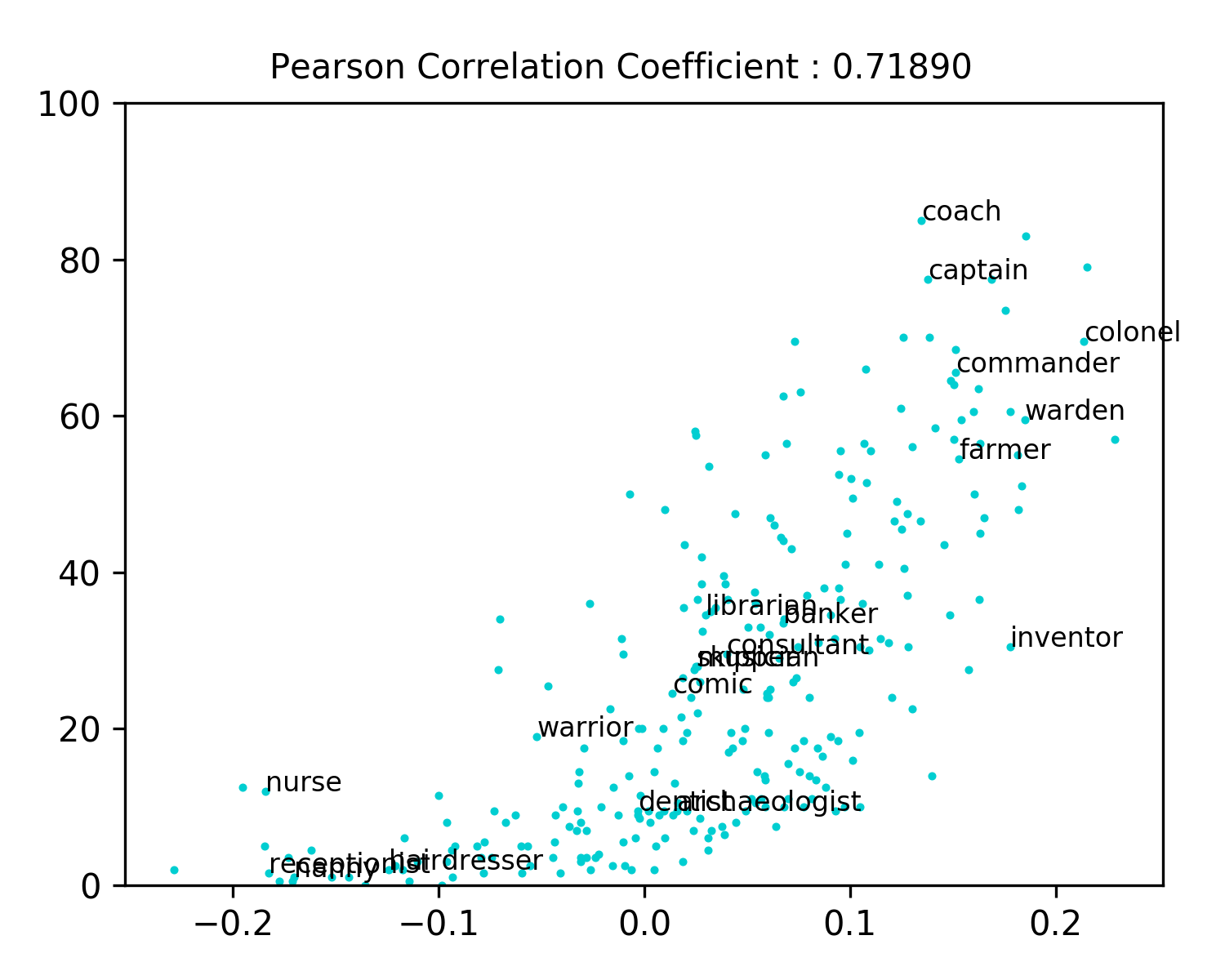} 
  \caption{ATT-Debias (0.7189)} 
  \end{subfigure}
  \begin{subfigure}{0.49\linewidth} 
  \centering
  \includegraphics[width=2.2in,height=1.2in]{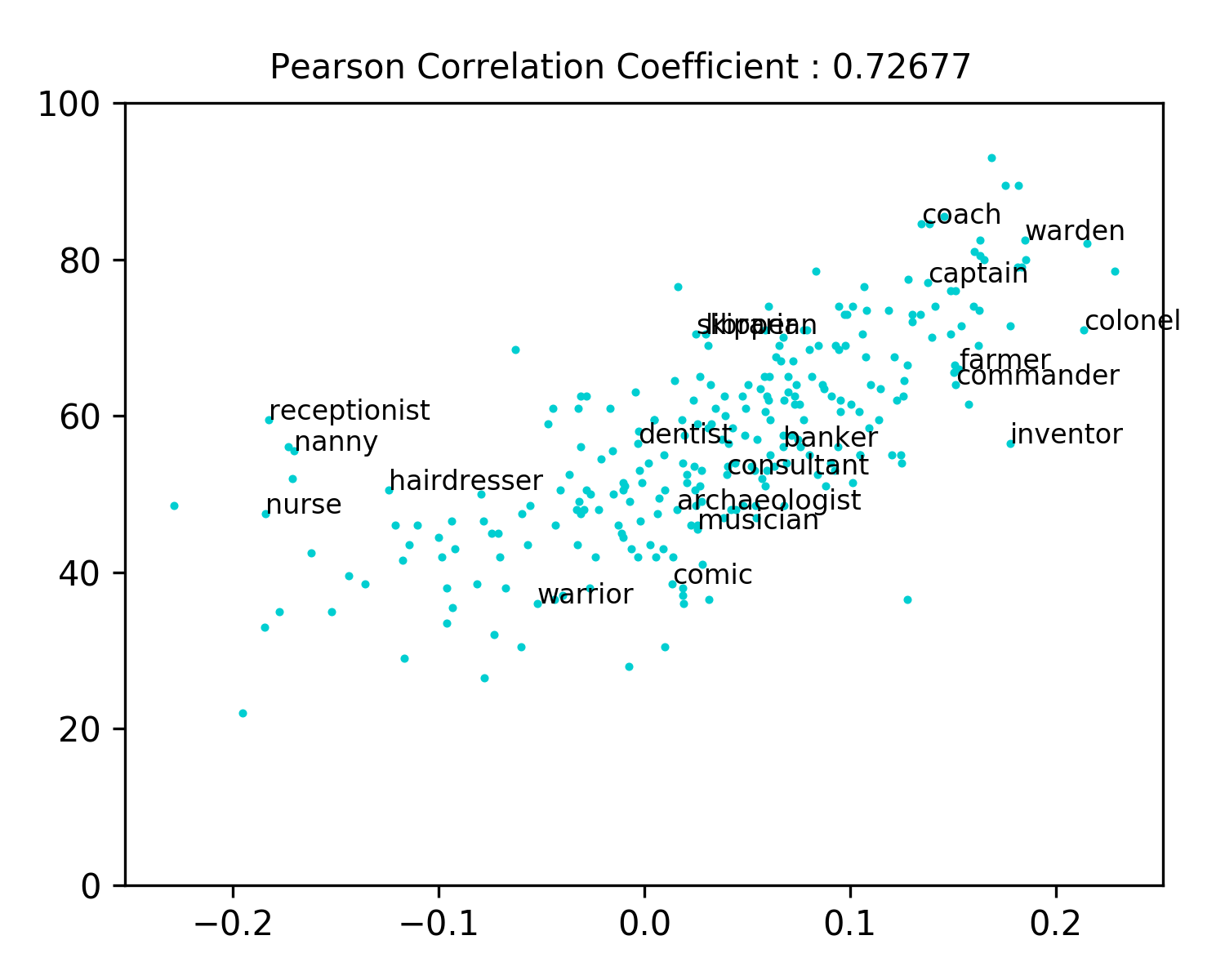} 
  \caption{CPT-Debias (0.7268)} 
  \end{subfigure}  
  \hfill 
  \begin{subfigure}{0.49\linewidth} 
  \centering
  \includegraphics[width=2.2in,height=1.2in]{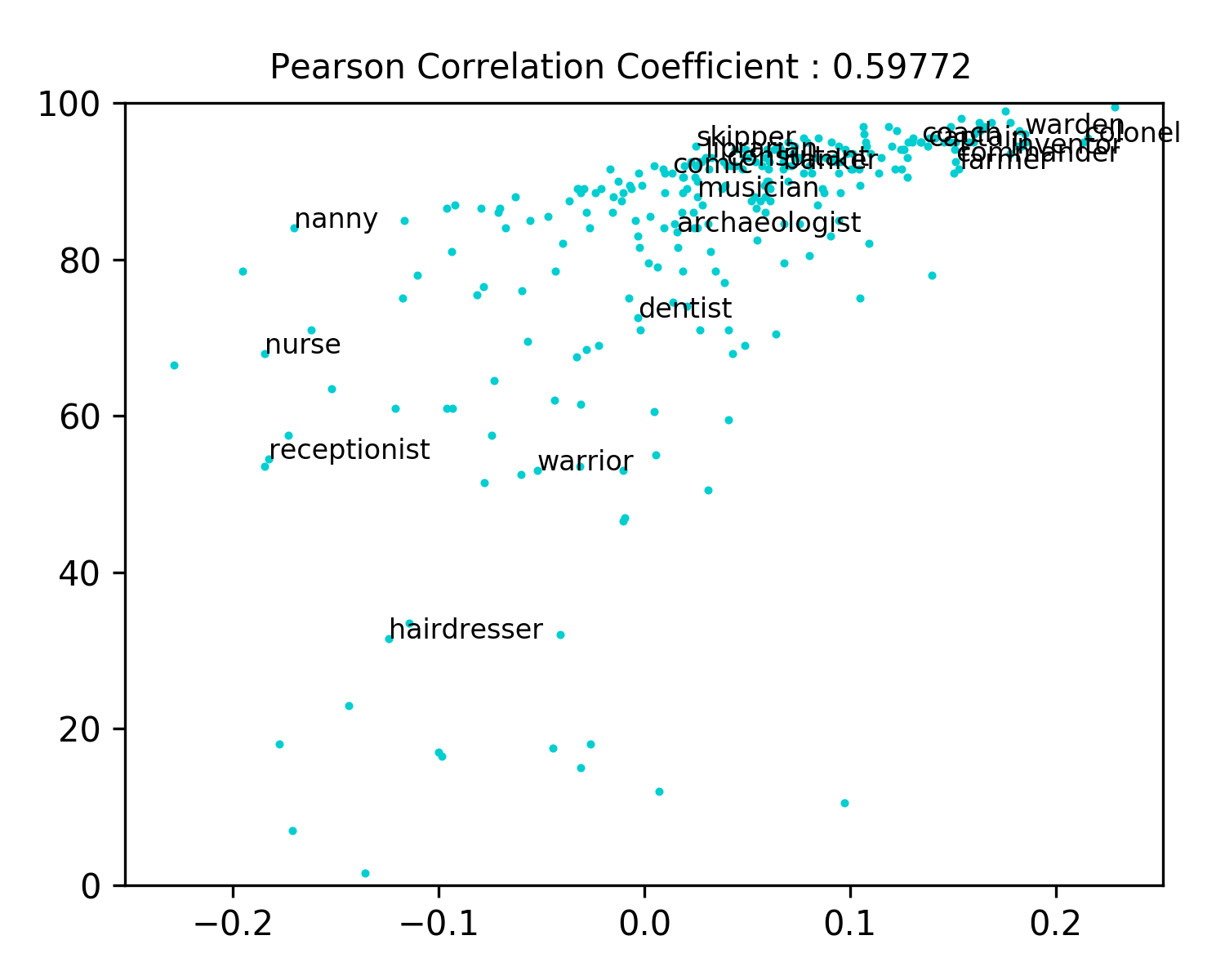} 
  \caption{AE-Debias (0.5977)} 
  \end{subfigure}
  \begin{subfigure}{0.49\linewidth} 
  \centering
  \includegraphics[width=2.2in,height=1.2in]{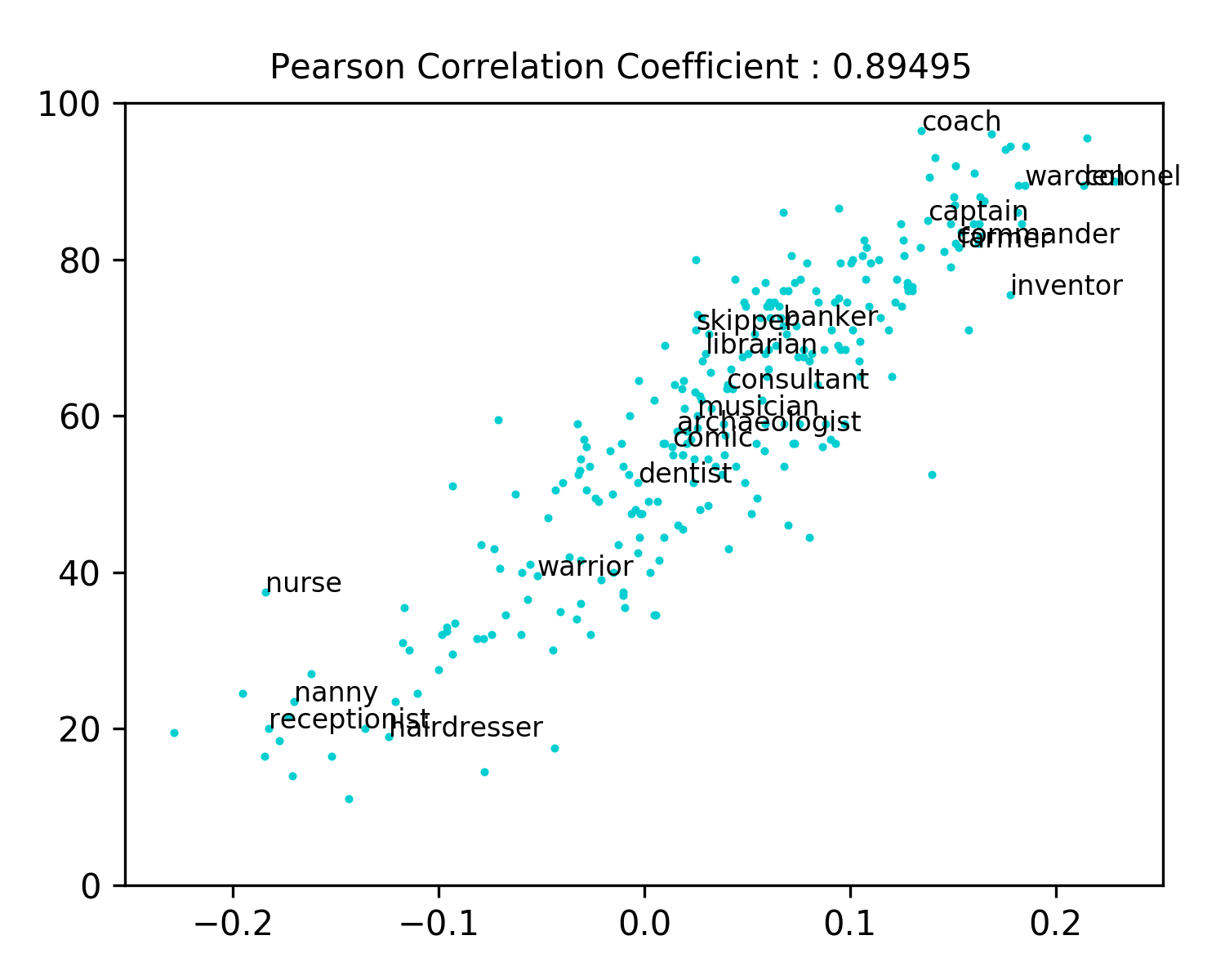} 
  \caption{AE-GN-Debias (0.8950)} 
  \end{subfigure}  
  \hfill 
  \begin{subfigure}{0.49\linewidth} 
  \centering
  \includegraphics[width=2.2in,height=1.2in]{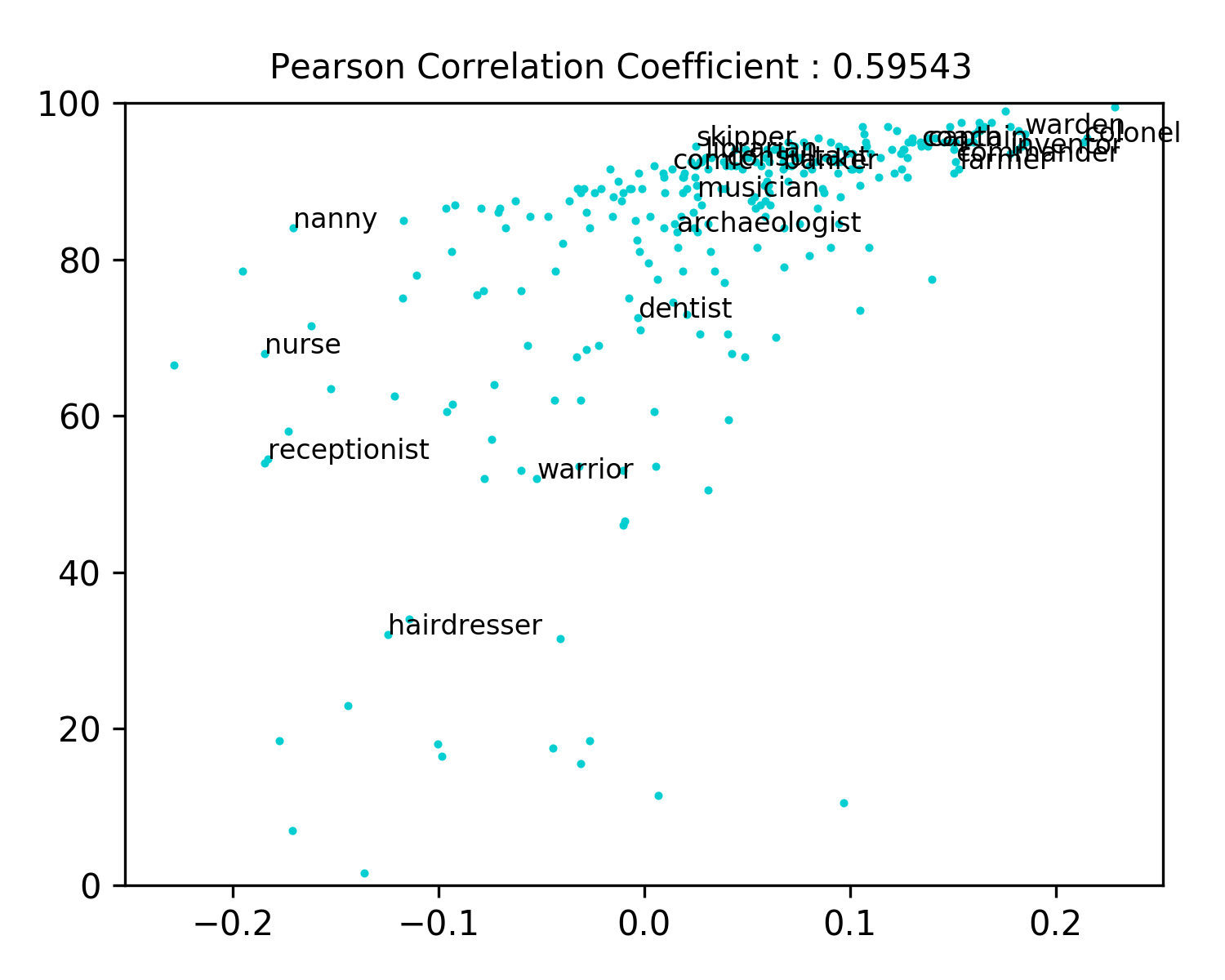} 
  \caption{GP-Debias (0.5954)} 
  \end{subfigure}
  \begin{subfigure}{0.49\linewidth} 
  \centering
  \includegraphics[width=2.2in,height=1.2in]{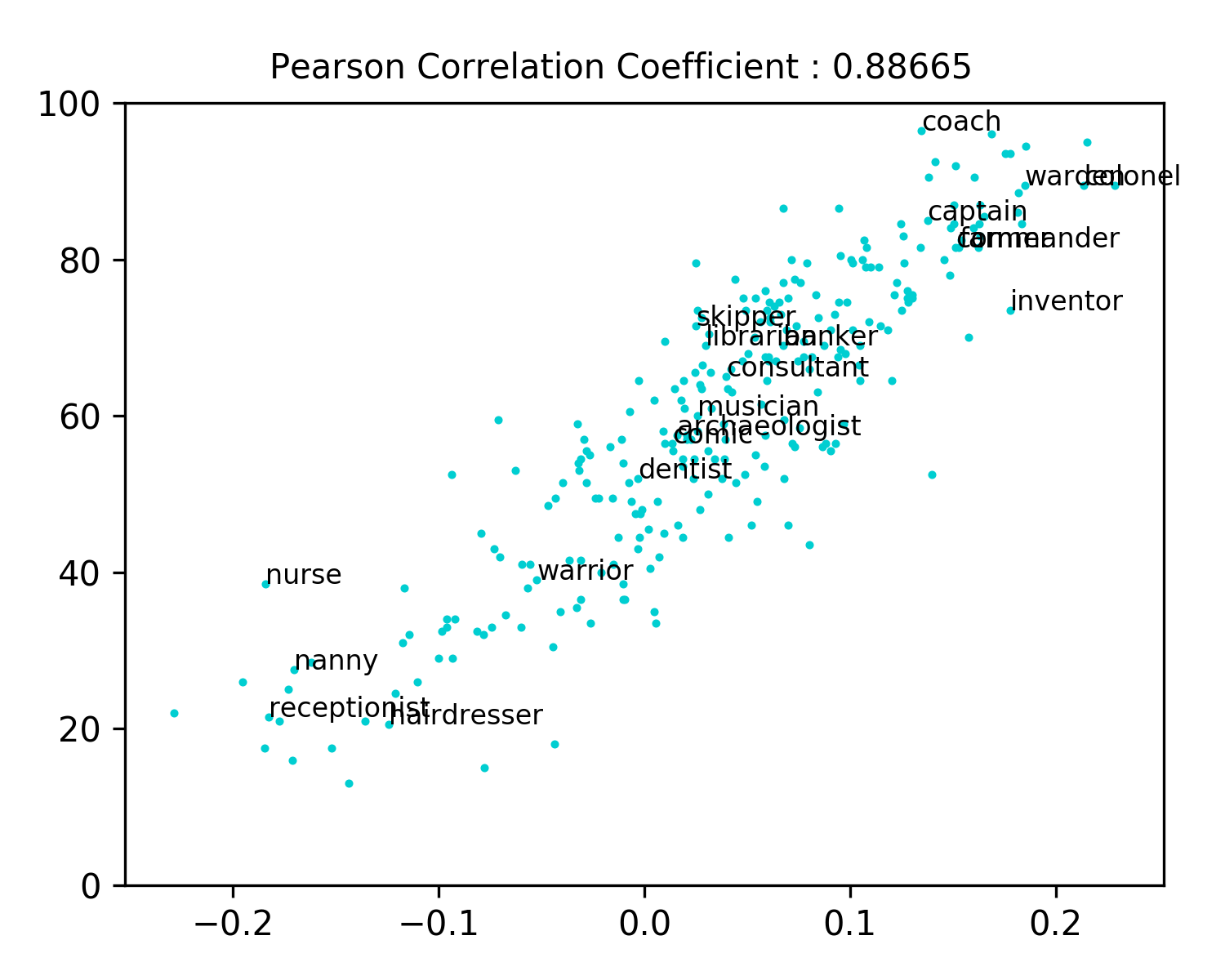} 
  \caption{GP-GN-Debias (0.8867)} 
  \end{subfigure}  
  \hfill 
  \begin{subfigure}{0.49\linewidth} 
  \centering
  \includegraphics[width=2.2in,height=1.2in]{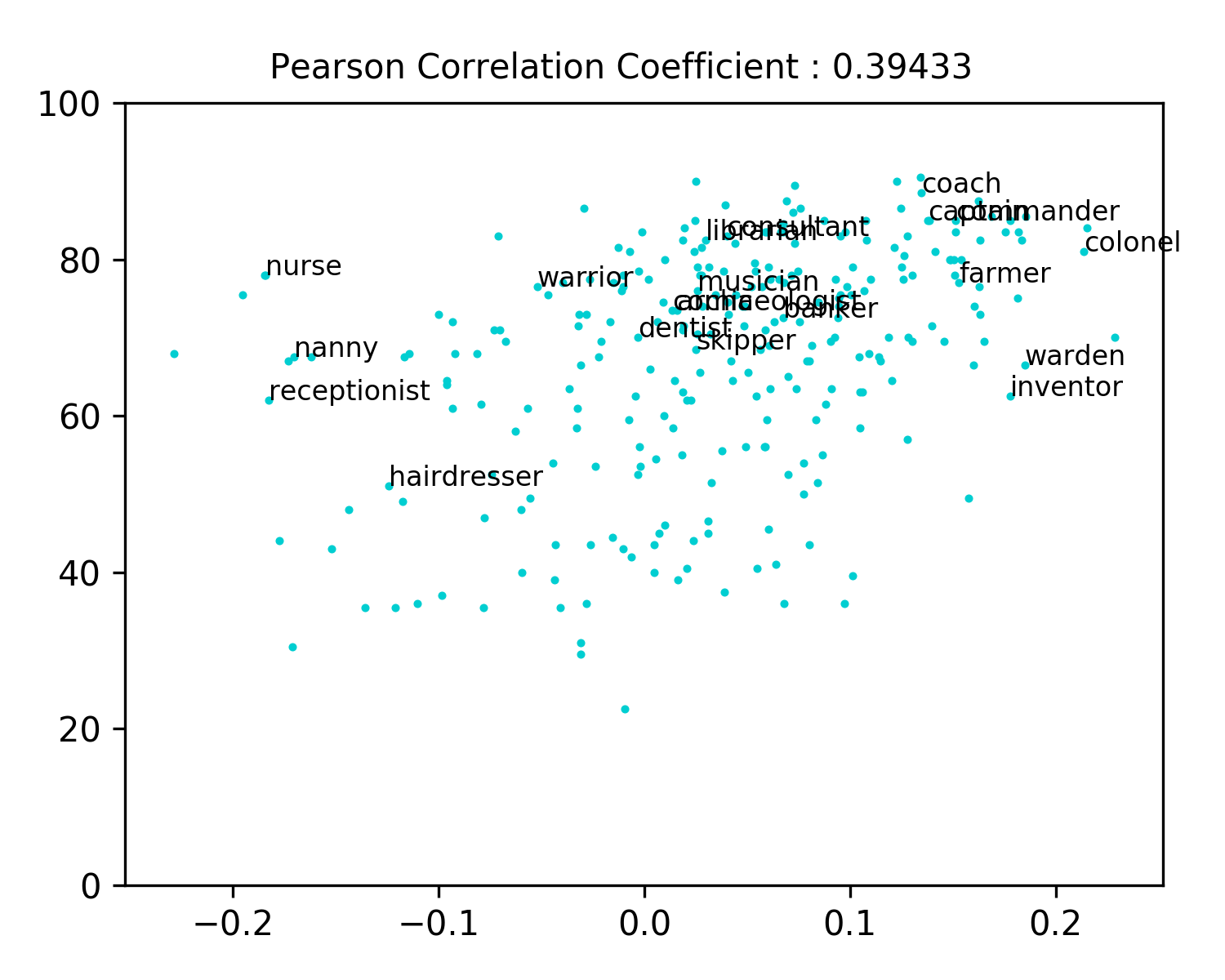} 
  \caption{CF-Debias (0.3943)} 
  \end{subfigure}
  \begin{subfigure}{0.49\linewidth} 
  \centering
  \includegraphics[width=2.2in,height=1.2in]{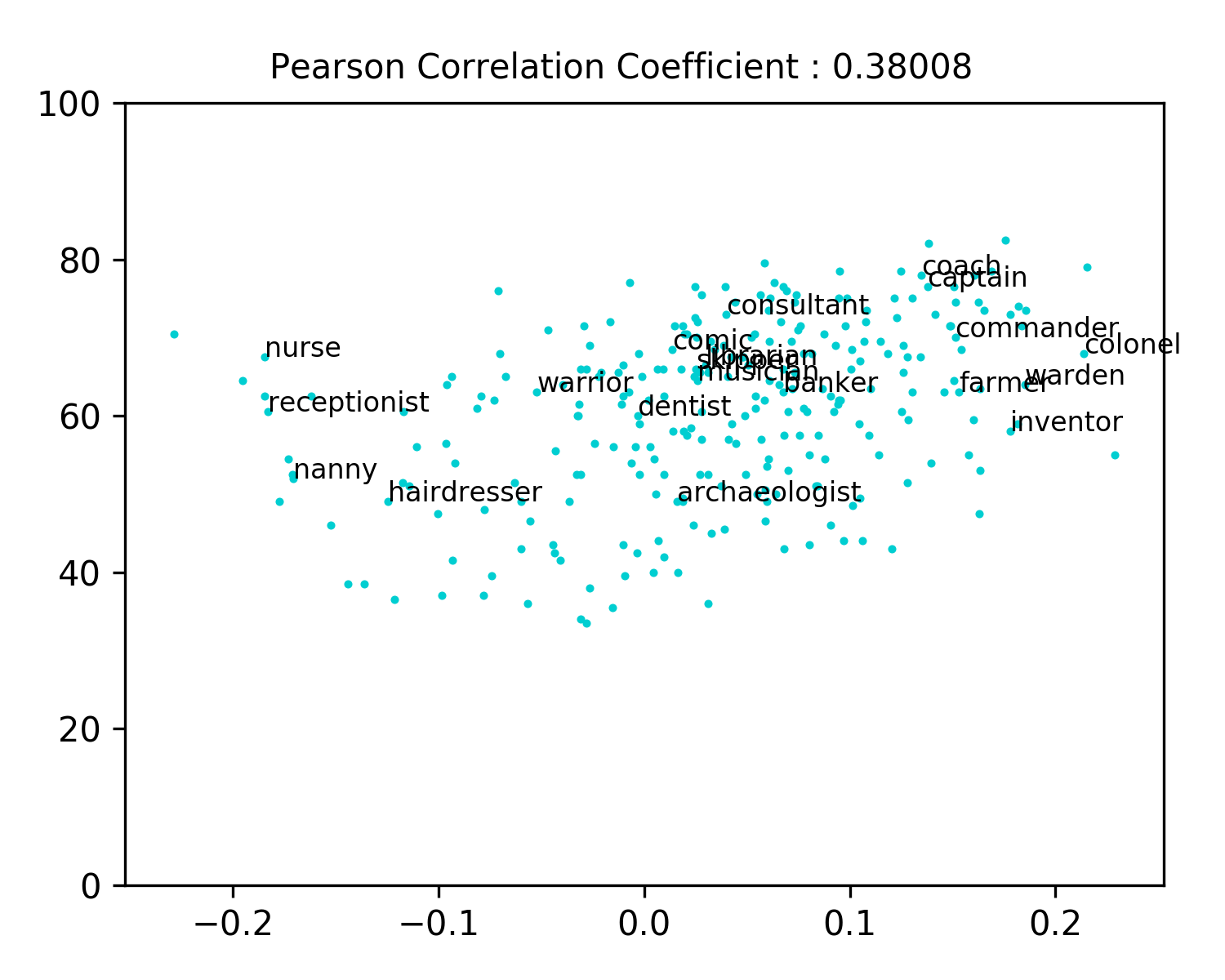} 
  \caption{CF-Debias-LA (0.3801)} 
  \end{subfigure}  
  \hfill 
  \begin{subfigure}{0.49\linewidth} 
  \centering
  \includegraphics[width=2.2in,height=1.2in]{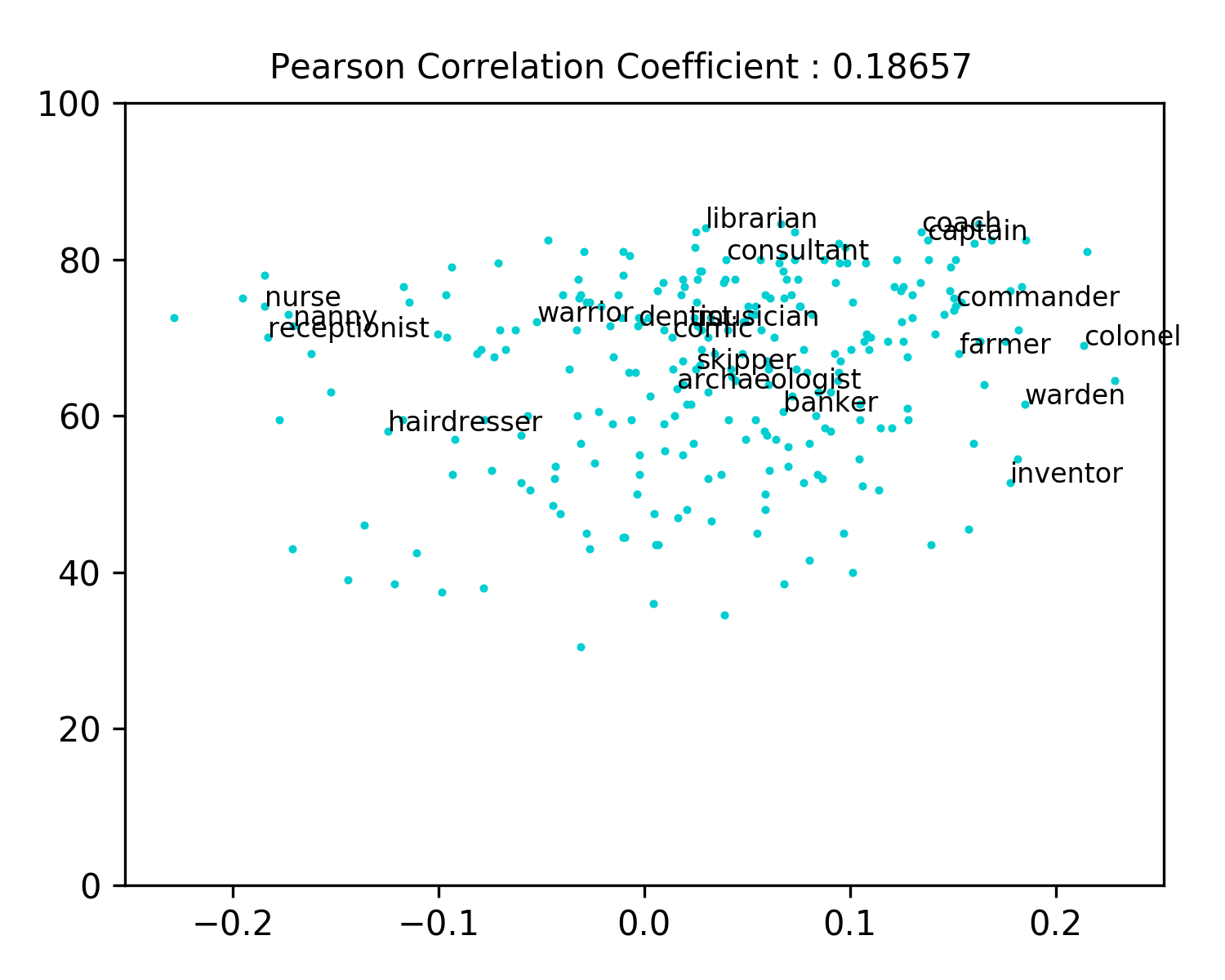} 
  \caption{CF-Debias-KA (0.1865)} 
  \end{subfigure}
  \caption{The percentage of male neighbors for each profession as a function of original bias for whole embeddings, we show only a limited number of professions on the plot to make it readable. The pearson correlation coefficient is added in parenthesis.}
  \label{fig:cluster}
  \end{figure}
\FloatBarrier

\newpage
\twocolumn
\section{Experimental Setup for Our Method}
\noindent We implement the encoder $E$ and the decoder $D$ with one hidden layer and hyperbolic tangent function as an activation function. The generators $C_{a}$ and $C_{g}$ are implemented as feed-forward neural network with one hidden layer, followed by the hyperbolic tangent function as an activation function. The gender classifier $C_{r}$ is similarly implemented as the feed-forward neural network with one hidden layer, followed by sigmoid activation function for the output layer. The whole training was performed using the Adam optimizer with learning rate $10^{-5}$. We trained our model using a single Titan-RTX GPU. Each run takes approximately 2 hours including the time for saving the post-processed word embeddings. 
\FloatBarrier
As described in Appendix \ref{sec:genclass}, to test classification accuracy of the gender classifier $C_r:\boldsymbol{z}^{g} \rightarrow [0,1]$ for gender-definition words and gender stereotypical words, we only used 143 gender word pairs from entire gender word pairs on the training procedure. The remaining 53 gender word pairs were utilized for gender classification test in Appendix \ref{sec:genclass}.

\section{The Link of Implementation for Each Baseline}
\noindent Hard-GloVe : \url{https://github.com/tolga-b/debiaswe}. \\
GN-GloVe : \url{https://github.com/uclanlp/gn_glove}. \\
CPT-GloVe : \url{https://github.com/jsedoc/ConceptorDebias}. \\
ATT-GloVe : \url{https://github.com/sunipa/Attenuating-Bias-in-Word-Vec}. \\
AE-GloVe, AE-GN, GP-GloVe and GP-GN : \url{https://github.com/kanekomasahiro/gp_debias}.

\section{The Experimental Setting of Human Experiment}
\noindent We conducted an human validation test on the linear analogies generated by the debiased embeddings to evaluate debiasing efficacy of each embedding. For the question "$a$ is to $b$ as $c$ is to ?", words $a,b$ were selected from gender word pairs of $Sembias$ dataset and $c$ was sampled from gender stereotypical words, i.e. homemaker, given by \citet{bolukbasi16}. \\
\indent The question word is chosen from ${argmax}_{d \in V}(\overrightarrow{d}\boldsymbol{\cdot} (\overrightarrow{c}{-}\overrightarrow{a}{+}\overrightarrow{b}))$. In order to enable human subjects to efficiently compare generated words of each debiased word embedding, We compared only 5 baseline methods; Original GloVe embedding, Hard-Debias, ATT-Debias, CPT-Debias, GP-GN-Debias with our methods; CF-Debias-LA and CF-Debias-KA. As stated in section 4.4 of main paper, 18 Human subjects were asked to evaluate the 84 generated analogies from two perspectives; 1) the existence of gender bias on generated analogy, 2) the semantic validity of analogy. The semantic validity in our experiment equals to the question, "Is it possible to infer semantic relationship from generated analogy?". The representative examples of the analogy questions are given as follows: "$man$ is to $woman$ as $boss$ is to ?" , "$female$ is to $male$ as $weak$ is to ?". 

\section{The Experimental Settings for Other Languages; Spanish and Korean}
\noindent We used Fasttext \cite{fasttext} pre-trained on $Common Crawl$ and $Wikipedia$, with 300 dimensional embeddings for 2,000,000 unique words for the experiments of Spanish. Also, we used Fasttext \cite{fasttext} pre-trained on $Wikipedia$, with 300 dimensional embeddings for 879,125 unique words for the experiments of Korean. For the gender word pairs required for gender debiasing, the query words used in the English version were translated into Spanish and Korean. In this procedure, some words, which are not present in the given corpus, were excluded.


\end{document}